\documentclass[sigconf, nonacm, pdfa]{acmart}

\usepackage{colorprofiles}
\usepackage[a-2b, mathxmp]{pdfx}
\usepackage{amsmath}
\usepackage{colortbl}
\usepackage{booktabs, tabularx}
\usepackage{xspace}

\let\oldtexttt\texttt
\renewcommand{\texttt}[1]{\oldtexttt{\small #1}}

\let\oldunderline\underline
\renewcommand{\underline}[1]{\oldunderline{\smash{#1}}}

\usepackage{enumitem}
\setlist[itemize]{noitemsep, topsep=0pt, leftmargin=1em}

\newcommand{\gcbuf}{wave buffer}

\newcommand{\dsAttIdx}{wave index}

\newcommand{\minisec}[1]{\noindent\textbf{#1. }}

\newcommand{\minisubsec}[1]{\noindent\underline{#1. }}

\usepackage{tikz}
\newcommand*\circleb[1]{\tikz[baseline=(char.base)]{
    \node[shape=circle,fill,inner sep=.5pt](char){\textcolor{white}{\small #1}};}}

\newcommand\vldbdoi{10.14778/3796195.3796212}
\newcommand\vldbpages{1016-1031}
\newcommand\vldbvolume{19}
\newcommand\vldbissue{5}
\newcommand\vldbyear{2026}

\newcommand\vldbtitle{\shorttitle} 
\newcommand\vldbavailabilityurl{https://github.com/microsoft/RetrievalAttention}
\newcommand\vldbpagestyle{empty}

\usepackage{tikz}

\newcommand{\newsysname}{\textsc{RetroInfer}\xspace}

\begin{document}

\renewcommand{\thefootnote}{\fnsymbol{footnote}}

\title{RetroInfer: A Vector Storage Engine for Scalable \\ Long-Context LLM Inference}


\author{
Yaoqi Chen$^{1,2*}$, 
Jinkai Zhang$^{3,2*}$, 
Baotong Lu$^{2 \dagger}$, 
Qianxi Zhang$^{2}$, 
Chengruidong Zhang$^{2}$, 
Jing Liu$^{2}$, 
\\Jingjia Luo$^{4,2*}$, 
Di Liu$^{5}$, 
Huiqiang Jiang$^{2}$, 
Qi Chen$^{2}$, 
Bailu Ding$^{2}$, 
Xiao Yan$^{3,6}$, 
Jiawei Jiang$^{3}$,
\\ Chen Chen$^{5}$, 
Mingxing Zhang$^{4}$, 
Cheng Li$^{1,7}$, 
Yuqing Yang$^{2}$,
Fan Yang$^{2}$, 
Mao Yang$^{2}$}
\affiliation{%
  \institution{$^{1}$University of Science and Technology of China \quad $^{2}$Microsoft Research \quad $^{3}$Wuhan University \\ $^{4}$Tsinghua University \quad 
  $^{5}$Shanghai Jiao Tong University  \quad $^{6}$Institute for Math \& AI, Wuhan \\ $^{7}$Institute of Artificial Intelligence, Hefei Comprehensive National Science Center
\\ 
{$^{2}$}\{baotonglu, qianxi.zhang, chengzhang, jingliu3, hjiang, cheqi, Yuqing.Yang, fanyang, maoyang\}@microsoft.com 
\\ $^{1}$yaoqi\_chen@mail.ustc.edu.cn, chengli7@ustc.edu.cn
$^{3}$\{zhangjinkai, yanxiaosunny, jiawei.jiang\}@whu.edu.cn
\\ $^{4}$\{luojj22, zhang\_mingxing\}@mail.tsinghua.edu.cn 
$^{5}$\{liu-di, chen-chen\}@sjtu.edu.cn $^{2}$bailuding@gmail.com}
}

\begin{abstract}

Recent large language models (LLMs) are rapidly extending their context windows,
yet inference throughput lags due to increasing GPU memory and bandwidth
demands. This is because the key-value (KV) cache, an intermediate structure
storing token representations, grows linearly with context length and requires an
iterative linear scan for attention computation. A promising direction to
accelerate long-context inference is to exploit attention's inherent sparsity by
offloading the KV cache to CPU memory and retrieving only a small subset of
tokens important to the current generation step. However, prior sparse attention
approaches struggle to balance accuracy and retrieval cost due to varying
sparsity patterns and inefficient GPU-CPU memory management.

We present \newsysname{}, a vector storage engine that realizes a sparsity-based
KV cache for long-context inference. \newsysname{} introduces an \underline{A}ttention-a\underline{W}are \underline{VE}ctor index (\textit{wave index}), 
which fundamentally improves the tradeoff between
attention accuracy and retrieval cost through tripartite attention
approximation, accuracy-bound attention estimation, and segmented clustering. 
We also design the \textit{wave buffer}, a GPU-CPU buffer manager that assigns
computation and manages data across heterogeneous hardware. We evaluate
\newsysname{} across a range of models and workloads, demonstrating up to
4.4$\times$ decoding throughput over full attention at 120K context and
up to 12.2$\times$ over sparse attention baselines at 1 million tokens—all while
preserving full-attention-level accuracy.
\end{abstract}

\maketitle

\footnotetext[1]{Work performed during the internship while at Microsoft.}
\footnotetext[2]{Corresponding author: Baotong Lu.}
\renewcommand{\thefootnote}{\arabic{footnote}}  

\pagestyle{\vldbpagestyle}
\begingroup\small\noindent\raggedright\textbf{PVLDB Reference Format:}\\
Yaoqi Chen, Jinkai Zhang, Baotong Lu, Qianxi Zhang, 
Chengruidong Zhang, Jing Liu, Jingjia Luo, Di Liu, Huiqiang Jiang, Qi Chen, Bailu Ding, Xiao Yan, 
Jiawei Jiang, Chen Chen, Mingxing Zhang, Cheng Li, Yuqing Yang, Fan Yang, Mao Yang. 
\vldbtitle. PVLDB, \vldbvolume(\vldbissue): \vldbpages, \vldbyear.\\
\href{https://doi.org/\vldbdoi}{doi:\vldbdoi}
\endgroup
\begingroup
\renewcommand\thefootnote{}\footnote{\noindent
This work is licensed under the Creative Commons BY-NC-ND 4.0 International License. Visit \url{https://creativecommons.org/licenses/by-nc-nd/4.0/} to view a copy of this license. For any use beyond those covered by this license, obtain permission by emailing \href{mailto:info@vldb.org}{info@vldb.org}. Copyright is held by the owner/author(s). Publication rights licensed to the VLDB Endowment. \\
\raggedright Proceedings of the VLDB Endowment, Vol. \vldbvolume, No. \vldbissue\ %
ISSN 2150-8097. \\
\href{https://doi.org/\vldbdoi}{doi:\vldbdoi} \\
}\addtocounter{footnote}{-1}\endgroup

\ifdefempty{\vldbavailabilityurl}{}{
\vspace{.3cm}
\begingroup\small\noindent\raggedright\textbf{PVLDB Artifact Availability:}\\
The source code, data, and/or other artifacts have been made available at \url{\vldbavailabilityurl}.
\endgroup
}

\section{Introduction}
\label{sec:intro}

Transformer-based LLMs that rely on attention
mechanisms~\cite{vaswani2017attention} have experienced rapid adoption and
enabled a range of new applications. Recently, their context windows have
expanded dramatically to support use cases like multi-turn
conversations~\cite{gao2024cost}, code and data
analysis~\cite{bairi2024codeplan, li2023can}, and reasoning~\cite{wei2022chain}.
Many popular LLMs now support context windows of 128K tokens~\cite{chatgpt,
claude}, and leading models such as Gemini Pro~\cite{gemini} and
Llama 4~\cite{llama4} push this further to 10 million tokens.

Scaling inference throughput with increasing context length presents
significant challenges to GPU resources.  During inference, the memory
consumption of an LLM's key-value (KV) cache~\cite{pope2023efficiently} -- a core component
for accelerating attention -- grows linearly with the sequence
length~\cite{kwon2023efficient}, straining memory capacity.
For instance, serving a single 1M-token request with Llama3-8B~\cite{llama3-8B-1048k} requires up to 125GB of memory.  
In addition, generating a new token requires iterating over all vectors in the
KV cache, resulting in excessive memory access that quickly drains GPU
bandwidth. 
Even with multiple GPUs, which offer more aggregate memory, large models with
large KV caches continue to stress both capacity and bandwidth.

A promising direction is to exploit the inherent sparsity of attention
to build a sparsity-based KV cache~\cite{child2019generating,correia2019adaptively,deng2024sparse}.
Recent systems~\cite{quest,liu2024retrievalattention,magicpig} show that a
small subset of tokens dominates attention output for a given query.  
Using only important tokens for attention calculation significantly reduces
memory access and enables offloading the KV cache to high-capacity, slower CPU
memory. However, identifying important tokens remains challenging.

Vector index~\cite{indyk1998approximate, ram2012maximum, shrivastava2014asymmetric, fu2017fast, malkov2018efficient, wang2020deltapq, Lu2021hvs, zhao2023towards},
traditionally used for approximate nearest neighbor search (ANNS), has
shown promise in identifying these important
tokens~\cite{liu2024retrievalattention, zhang2024pqcache, alayadb}.  ANNS seeks the most
similar vectors to a given query using inner product, which closely
aligns with how attention computes relevance (Section~\ref{sec:bg}).

\begin{figure}[t!]
    \centering
    \includegraphics[width=0.90\columnwidth]{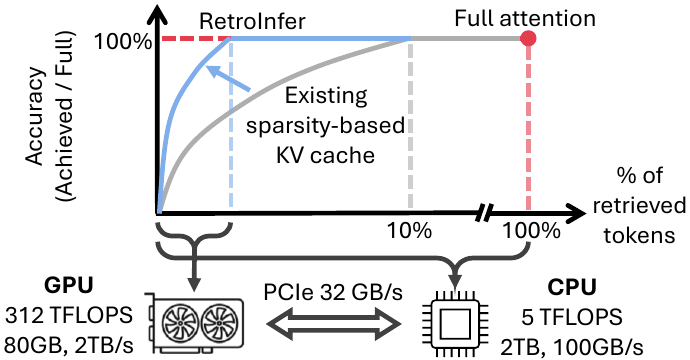}
    \caption{\newsysname raises the trade-off ceiling between accuracy and retrieval cost, and manages data across hardware.}
    \label{fig:tradeoff}
\end{figure}

Yet, sparsity-based KV cache systems face a fundamental trade-off between
attention accuracy and retrieval cost. Existing vector indexes fall short due to
the variable nature of attention sparsity. Important tokens differ across
layers, attention heads, decoding steps, and input tasks~\cite{magicpig,
Pyramid}; thus, standard indexes struggle to retrieve a small number of tokens
while maintaining high accuracy. As shown in Figure~\ref{fig:tradeoff}, current
systems reduce data volume but still incur high retrieval costs (e.g., 10\%) to
preserve accuracy. We argue that a holistic, attention-aware approach for KV
cache is needed to support long-context inference effectively.

We present \newsysname{}, a vector storage engine that holistically addresses
the challenges of sparsity-based KV cache systems.
Central to \newsysname{} are two components: 
an \textit{Attention-aWare VEctor index} (wave index) and an
\textit{accuracy-agnostic GPU-CPU buffer manager} (wave buffer).
Our design follows two
principles: (1) be attention-aware, and (2) separate the concern of attention
accuracy from the system efficiency.

Wave index is a new vector index design that improves the trade-off between
attention accuracy and retrieval cost.
Drawing inspiration from the classic vector indexes~\cite{sivic2003video}, wave
index introduces three novel attention-aware techniques. 
First, a {\textit{tripartite attention approximation}} logically partitions
tokens into steady, retrieval, and estimation zones; token importance decreases
across these zones. The steady zone computes precise attention on a small number
of consistently important tokens, the retrieval zone uses index-guided selection
to compute attention precisely, and the estimation zone approximates attention
for the remaining tokens based on the index structure.  As we move from steady
to estimation, attention becomes more approximate but cheaper to compute,
reducing sensitivity to the number of tokens retrieved and
enabling control over compute
cost without sacrificing accuracy.

Second, an
{\textit{accuracy-bound attention estimation mechanism}} approximates the
contribution of less important but non-negligible tokens with guaranteed
accuracy bounds in the estimation zone.  Third, a {\textit{segmented
clustering}} redesigns traditional vector index clustering for parallelism and
low overhead, reducing both the index construction and update cost.

With wave index, the key remaining challenge is to distribute computation -- index traversal,
construction, and attention calculation -- and to coordinate data movement
between CPU and GPU memory. 
The inherent ratio of important tokens is not low enough to fully hide PCIe
transfer latency, which can stall GPU computation.  The asymmetry between CPU
and GPU leads to mismatches between computation type and available resources. As
shown in Figure~\ref{fig:tradeoff}, the GPU provides high compute throughput but
has limited memory capacity, while the CPU offers much larger memory but slower
floating-point performance and memory access. The CPU is also better suited for
a broader range of computation. Great care must be taken to fully utilize
available resources, avoid GPU stalls, control software overhead, and reduce
PCIe bandwidth contention.

We design the wave buffer to address these challenges; it acts as the metadata
and control plane, similar to a buffer manager in a database
system~\cite{ziegler2022scalestore, leis2023virtual, leis2024leanstore}. 
We observe that access to important tokens exhibits strong temporal locality, so
the wave buffer incorporates a GPU-side block cache of the KV cache to further
reduce PCIe pressure. Attention calculation and block cache access are
performed synchronously on the GPU, while block cache replacement is handled
asynchronously. The wave buffer overlaps computation, reduces overhead, and
sustains overall inference efficiency.
The wave buffer is designed to be accuracy-agnostic, focusing solely on hardware
efficiency without affecting attention accuracy; it leverages classic database
techniques such as caching and pipelining.

We evaluate \newsysname{} on four popular models across a range of tasks
from advanced long-context benchmarks (e.g., 
RULER~\cite{hsieh2024ruler}) with varying context lengths and batch sizes. 
We also evaluate \newsysname{} on two reasoning models that generate long outputs.
Our experiments show significant 
improvements in inference efficiency while maintaining the accuracy of full attention.
For the same retrieval budget (i.e., the number of KVs retrieved), 
\newsysname{} outperforms existing sparsity-based baselines by 1.40--46.47\% in 
task accuracy and is the only system that
achieves accuracy comparable to full attention.
\newsysname{} also significantly improves inference throughput 
by supporting large batch sizes and context lengths.
Specifically, \newsysname{} achieves up to 4.4$\times$ throughput over full attention 
when the context length fits in GPU memory, and up to 12.2$\times$ throughput over 
other sparse-attention systems when the KV cache is extended to CPU memory. 
For end-to-end comparison, \newsysname achieves 2.2$\times$--3.3$\times$ throughput compared to vLLM.
We make the following contributions:
\begin{itemize}
    \item We present a novel vector index design -- wave index -- that improves the trade-off
    between attention accuracy and retrieval cost, enabling efficient and accurate
    long-context inference. Wave index introduces three novel techniques: tripartite
    attention approximation, accuracy-bound attention estimation, and segmented
    clustering.
    \item We introduce wave buffer, an accuracy-agnostic GPU-CPU buffer manager that
    efficiently manages data movement and computation across heterogeneous
    hardware.
    \item We design and implement \newsysname{}, a long-context inference system that
    integrates wave index and wave buffer. We evaluate \newsysname{} thoroughly on popular
    LLMs and tasks, demonstrating both high throughput and high
    accuracy.
\end{itemize}


\section{Background and Motivation}
\label{sec:bg}

\subsection{Transformer-based LLMs and Attention} 
\label{subsec:bg:llm}
\begin{figure}[t!]
    \centering
    \includegraphics[width=.96\linewidth]{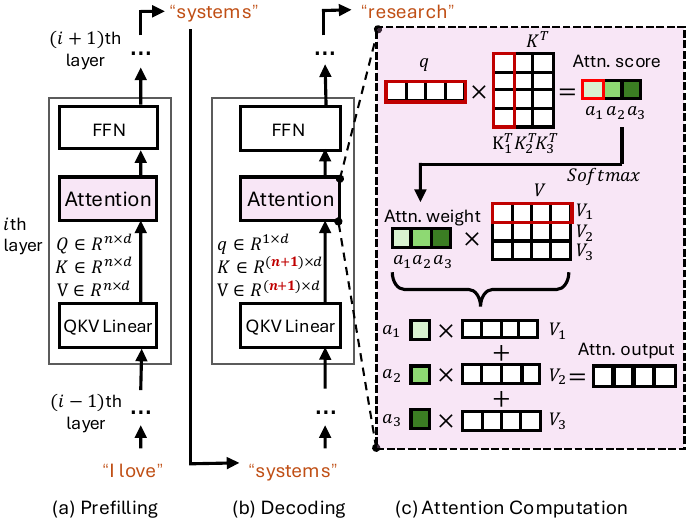}
    \caption{Transformer architecture with KV cache. In (c), deeper colors indicate higher attention weights (more important tokens). ``I love systems'' are the three input tokens.}
    \label{fig:full_attn}
\end{figure}

Transformers~\cite{vaswani2017attention} are the dominant architecture for large language models (LLMs), relying on a multi-layer attention mechanism. As shown
in Figure~\ref{fig:full_attn}, given an input of $n$ tokens, 
each represented as a high-dimensional vector, it is linearly transformed 
into three matrices: queries ($Q$), keys ($K$), and values ($V$). 
Within each layer's attention module, 
the model relates each token to others using multiple attention heads in parallel.
The outputs from all heads are then aggregated and passed to a feed-forward
neural network (FFN) layer. 
This process is repeated for each layer of the model.

During inference, the transformer operates in two distinct phases:
prefilling and decoding. In the prefilling phase, the model
processes the input prompt as a whole; in the decoding phase, it generates
tokens one at a time based on both the prompt and previously generated tokens.

To accelerate decoding, modern LLMs use a \textit{key-value cache} (KV cache)~\cite{pope2023efficiently, kwon2023efficient},
which stores the key and value vectors of all previously seen tokens. This
avoids redundant computation and significantly reduces decoding latency.
However, the KV cache introduces substantial memory and bandwidth requirements,
since its size and memory access grow linearly with both context length and batch size. 
Thus, KV cache makes long-context inference increasingly 
constrained by memory capacity and bandwidth limitations.

The attention output in decoding is computed as a weighted sum over value
vectors, where the weights are given by the softmax of the inner product between
the current query $q$ and all cached keys:
\begin{equation}
    \mathbf o = \mathbf a  \mathbf V = \text{Softmax}(\dfrac{\mathbf q \mathbf K^T}{\sqrt{d}})\mathbf{V} = \sum_{i=1..n} (\dfrac{
            e^{ \mathbf{q} \cdot \mathbf{K}^T_i / \sqrt{d} }
        }{
            \sum_{j=1..n}{ e^{ \mathbf{q} \cdot \mathbf{K}^T_j / \sqrt{d} } }
        }  \cdot \mathbf{V}_i)  
    \label{eq:attn}
\end{equation}

KV cache has become so essential that it has influenced model architecture changes in
modern LLMs. Notably, Grouped Query Attention (GQA)~\cite{ainslie2023gqa} has become
a standard in recent models~\cite{llama3.1,qwen2.5,yi6b,yi9b}.
In GQA, several query heads within a group share a key-value head, 
which reduces KV cache memory consumption proportionally to the group size.
However, the memory consumption remains substantial, especially in the 
long-context scenarios.

\subsection{Limits to Scaling Long-Context Inference}
Despite growing demand for long-context inference in use cases such as
multi-turn conversations~\cite{gao2024cost}, code understanding~\cite{bairi2024codeplan}, and
reasoning~\cite{wei2022chain}, achieving
scalable performance is challenging. While the KV cache reduces computation
cost, the widening gap between GPU memory capacity and memory requirements of KV cache severely
limits batch size and throughput.

\minisec{GPU Memory Capacity Limits}
The KV cache size grows linearly with both the context length and the
batch size, quickly draining GPU memory. For instance, 
an A100 GPU (80GB memory)~\cite{A100} can support a
maximum batch size of 4 at a 128K context length for Llama3-8B. 
Exceeding this limit results in out-of-memory (OOM) errors. 
Similarly, the maximum context length the GPU can support is 512K\footnote{Unless specified, the analysis is done in Llama3-8B-1048K~\cite{llama3-8B-1048k} on an A100 GPU. 
Experiments on more models and multiple GPUs are available in Section~\ref{sec:eval}.}.
Leveraging the aggregated memory from multiple GPUs is plausible but less cost-effective.
Thus, many systems offload the KV cache to larger and cheaper CPU memory~\cite{lee2024InfiniGen,sheng2023flexgen,xiao2024infllm}, but this incurs overhead due to PCIe transfers between CPU and GPU.

\minisec{GPU Memory Bandwidth Bottlenecks}
Memory bandwidth is another key limiting factor.  With KV cache, each query in
decoding phase linearly accesses all previous KV vectors.  As the context length
and batch size increase, the resulting access volume saturates memory bandwidth,
a limitation widely recognized in recent
work~\cite{chen2023accelerating,ribarsparq}.  In our experiments, scaling the
batch size beyond 3 (at 128K context) leads to marginal throughput improvement 
due to GPU memory bandwidth saturation.

\begin{figure}[t!]
    \centering
    \includegraphics[width=.92\linewidth]{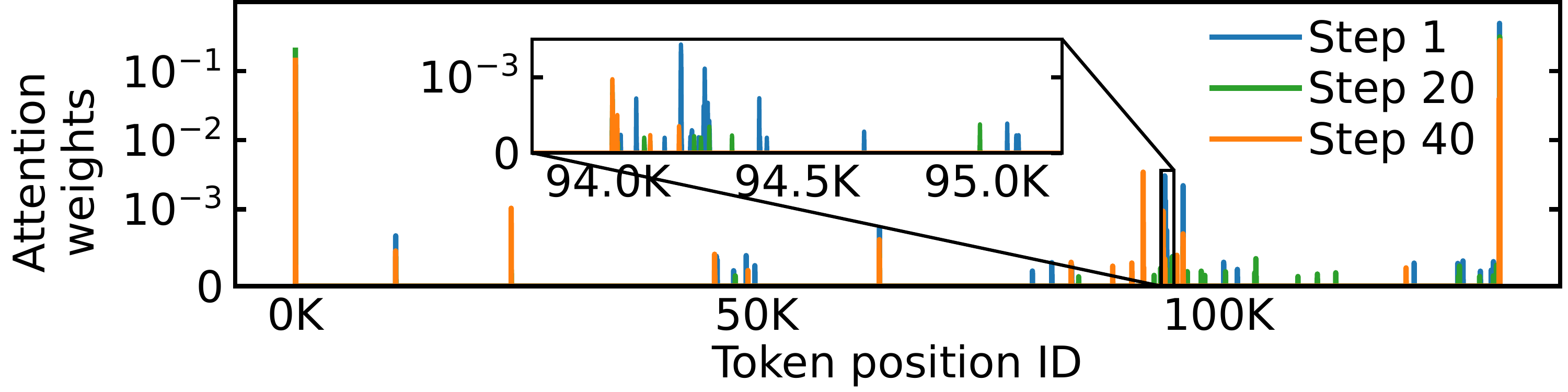}
    \caption{Dynamic sparsity in attention (Llama3-8B-1048K, \textit{qa\_1} task~\cite{hsieh2024ruler}, layer 0, head 24). The distribution of top-100 attention weights varies across decoding steps.}
    \label{fig:aivi_dist}
\end{figure}

\begin{figure}[t!]
    \centering
    \includegraphics[width=.91\linewidth]{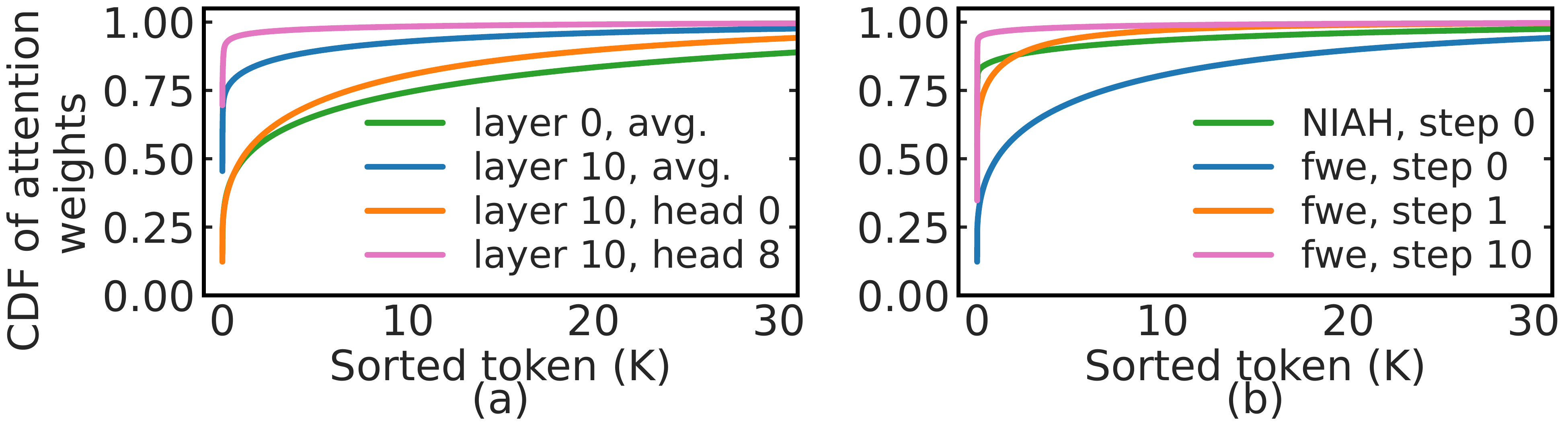}
    \caption{Variety of attention sparsity across (a) model layers and attention heads on \textit{fwe} task~\cite{hsieh2024ruler} and (b) decoding steps and tasks on Llama3-8B-1048K.}
    \label{fig:sparsity}
\end{figure}

\subsection{Sparsity-based KV Cache}
\label{subsec:bg:sparsity}
A promising direction to alleviate the memory and bandwidth bottlenecks of
the KV cache is to selectively access KV vectors, enabled by the inherent
\textit{sparsity} of the attention mechanism~\cite{child2019generating,correia2019adaptively,deng2024sparse}.
This sparsity arises naturally from the structure of the attention equation
(Equation~\ref{eq:attn}) and the exponential nature of the softmax function.
Intuitively, value vectors associated with high attention weights\footnote{In this paper, we refer to $\mathbf q \mathbf K^T$ in attention as the attention scores, $\text{Softmax}(\mathbf q \mathbf K^T / \sqrt{d})$ as the attention weights.}  
(i.e., elements of $\mathbf{a}$ in Figure~\ref{fig:full_attn}) 
dominate the attention output $\mathbf{o}$. That is,
a small subset of value vectors suffices to reconstruct the attention
output, provided that the most important tokens can be accurately identified from the vast pool of KV vectors. 

Leveraging sparsity decreases the GPU computation, reduces memory access and
helps alleviate the bandwidth bottleneck. Moreover, sparsity makes it feasible
to offload the KV cache to 
CPU memory to address the capacity limitation, as the inference system can 
selectively access a subset of KV vectors over PCIe. 

\minisec{Challenges to Leverage Sparsity}
Identifying important tokens is challenging for three reasons.
First, as shown in Figure~\ref{fig:aivi_dist}, important tokens (i.e., those
with high attention weights) are scattered across the context, making their
positions unpredictable. 
Second, tokens that are important at one decoding step (i.e., one query vector) may not
remain so in subsequent steps. As shown in Figure~\ref{fig:aivi_dist},
the overlap of the top-100 KV vectors across three decoding steps is
only 31\%, highlighting token importance dynamics.
Third, the sparsity exhibits high variability from two sources: the architecture of the model itself
and the nature of the decoding query. As shown in Figure~\ref{fig:sparsity},
attention distributions differ significantly across model layers and attention
heads (Figure~\ref{fig:sparsity}(a)), potentially reflecting the varying
semantic roles~\cite{retrieval_head,li2024snapkv}. 
Different queries within the same context or across different tasks exhibit
substantially different sparsity ratios on the same attention head (Figure~\ref{fig:sparsity}(b)).

Sparsity alone is not a panacea. A gap exists between sparsity ratio (i.e.,
negligible tokens percentage) and the bandwidth constraints imposed by the
PCIe interconnect. GPU memory bandwidth (e.g., HBM) is about 60$\times$
higher than PCIe bandwidth on common GPUs (e.g., A100, H100, and H200). To hide PCIe transfer latency and prevent GPU stalls, the sparsity ratio must
exceed 98\% for the GPUs.  Recent studies report lower sparsity ratios of 87.5\%~\cite{quest}
and 90\%~\cite{lee2024InfiniGen}, highlighting the need for careful system design that accounts for
hardware constraints to achieve high throughput.

\minisec{Existing Sparsity-based KV Cache Systems} 
Recent systems~\cite{streamingllm,li2024snapkv,xiao2024infllm,quest,lee2024InfiniGen,magicpig}
have explored the use of sparsity to facilitate long-context inference. However,
they often fall short along two key axes: accuracy and efficiency.

Some systems rely on fixed-position heuristics to discard KV
vectors~\cite{streamingllm,li2024snapkv}, often resulting in significant
accuracy loss due to static assumptions about token importance. Others estimate
importance by partitioning the KV cache into equal-sized chunks and using
representative vectors~\cite{quest,xiao2024infllm}. While GPU-friendly, this
coarse-grained approximation reduces inference accuracy.

Systems that leverage sparsity to reduce attention computation but retain all KV
vectors in GPU memory~\cite{quest} are constrained by GPU capacity, limiting
both context length and batch size.  Works such as
InfiniGen~\cite{lee2024InfiniGen} and MagicPIG~\cite{magicpig} offload the KV
cache to CPU memory and selectively fetch vectors, either via speculative
prediction or locality-sensitive hashing (LSH). However, inaccurate selection
can degrade quality, and throughput is bottlenecked by PCIe bandwidth or lower
CPU compute power.

\section{Vector Index for Sparsity-based KV Cache}
\label{sec:anns}

As pointed out in recent work~\cite{liu2024retrievalattention,zhang2024pqcache,alayadb},
vector index -- a classic technique for approximate nearest neighbor search
(ANNS)~\cite{indyk1998approximate, fu2017fast, malkov2018efficient, wang2020deltapq, Lu2021hvs, zhao2023towards} -- is a natural fit for retrieving important tokens and serves as the
backbone of a sparsity-based KV cache.  However, directly applying vector index to the
KV cache is not sufficient and introduces several challenges.

\minisec{Opportunities for Vector Index}
The key observation is that Maximum Inner Product Search (MIPS)~\cite{ram2012maximum, shrivastava2014asymmetric}, a variant of
nearest neighbor search (NNS), can be seamlessly applied to identify critical
tokens in attention
mechanisms~\cite{liu2024retrievalattention,zhang2024pqcache}. High attention
scores (i.e., important tokens) indicate that their key vectors have a large
inner product with the query vector (i.e., they are similar).

ANNS aims to identify a desired number of vectors most similar
to a given query in high-dimensional space, using a similarity metric such as
inner product. Vector index~\cite{fu2017fast, malkov2018efficient, wang2020deltapq, Lu2021hvs, zhao2023towards} is a
widely used technique for ANNS that organizes key vectors so
that similar vectors are placed nearby; it greatly reduces the number of
vectors accessed during search while maintaining near-optimal results.

Vector index is promising because it naturally handles the dynamic nature of
sparsity (Figure~\ref{fig:aivi_dist}): each decoding token (i.e., query vector) retrieves different results from the
index based on importance, rather than relying on fixed positions in the
context.

\minisec{Challenges}
The fundamental trade-off between accuracy and retrieval cost in ANNS persists
and becomes more pronounced when applied to sparsity-based KV caching. Existing
vector indexes are inadequate to address the accuracy challenge due to the high
variability in attention sparsity (Figure~\ref{fig:sparsity}). As in prior
work~\cite{zhang2024pqcache}, the retrieval cost
remains substantial relative to the limited PCIe bandwidth, because it must account
for this variability to retrieve more tokens for desired accuracy.

Moreover, efficiency challenges arise when using a vector index for
sparsity-based KV cache, as it introduces index traversal and selective data
access into an inference system optimized for dense GPU execution. A careful
co-design of compute and memory across the hardware stack is essential.

First, sparse attention with vector indexing involves two types of data (index
structures and KV vectors) and three types of computation (index traversal,
index construction, and attention). Mapping this onto hardware introduces
fundamental challenges due to CPU-GPU asymmetry. GPUs offer high-throughput
compute and fast memory access but have limited capacity; CPUs provide abundant
memory but lower compute throughput. This raises key questions about data
placement, task assignment, and how to leverage parallelism and overlap to
utilize both processors efficiently.

A natural starting point is to perform attention calculation on the GPU while storing
high-volume KV vectors in CPU memory. Under this setup, efficient data movement
is critical. Without careful orchestration, the GPU can be starved for data,
especially under tight latency constraints.

Second, the system must reduce index traversal and construction costs while
preserving retrieval quality. Index traversal can be costly on GPU due to
irregular memory access and fine-grained operations (e.g., top-$k$ selection in ANNS) 
while CPU-based traversal is constrained by weak compute power. 
Index construction adds additional overhead, and prior
work~\cite{alayadb,liu2024retrievalattention,hooper2024squeezed} assumes offline construction,
limiting the usability in online inference.

Finally, the control logic for managing index structures is often better suited
for CPU execution. This necessitates a memory management layer that runs
on the CPU and coordinates efficient paging and data transfers to the GPU. These
challenges are further complicated by the distinct programming models of CPUs
and GPUs, requiring careful coordination to ensure high performance and low
overhead.

\section{\newsysname{}}
\label{sec:design}

We present \newsysname, a vector storage engine that builds a sparsity-based KV cache to scale long-context inference.
We begin with a high-level overview, then describe the design of the \textit{wave index} and \textit{wave buffer}, 
and also discuss the implementation.

\begin{figure}[t!]
    \centering
    \includegraphics[width=\linewidth]{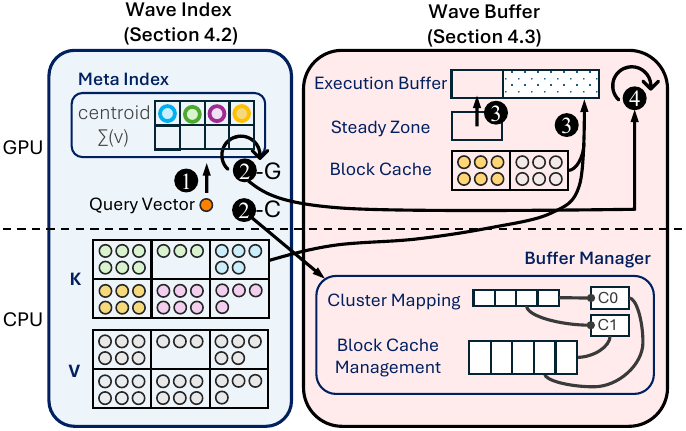}
    \caption{Architecture of \newsysname.
    Circles with numbers represent the steps of attention computation, and steps with the same number occur in parallel.
    Centroids (bold-edged circles) are stored in the meta index.
    }
    \label{fig:arch}
\end{figure}

\subsection{Overview}
The design of \newsysname{} follows two principles: (1) be attention-aware, and
(2) separate the concern of attention accuracy from the system efficiency.

We introduce an \underline{A}ttention-a\underline{W}are \underline{VE}ctor 
index called \textit{\dsAttIdx} that retrieves the important KV vectors accurately
and efficiently. We also introduce a \textit{\gcbuf}, which manages
memory across GPU and CPU to facilitate \dsAttIdx, also in an attention-aware manner.
Figure~\ref{fig:arch} shows the architecture of \newsysname{}.

To make judicious use of the GPU memory, the wave index employs a cluster-based
vector index design.  Specifically, it partitions KV vectors into
clusters based on their similarity, storing cluster centroids in a
{\textit{meta index}} as their representatives in GPU memory. 
Each meta index entry contains additional information (i.e., $\sum V$),
which allows us to perform the precise attention computation on retrieved clusters,
while approximating attention for others to cover varying sparsity ratios.
All of the KV vectors are organized in contiguous KV blocks in CPU
memory. We dive into the internals of wave index in Section~\ref{subsec:wave-index}.

The wave buffer serves two purposes. First, it contains several buffers in GPU
memory to accelerate inference throughput. These include a {\textit{block cache}} for KV
vectors and an {\textit{execution buffer}}, a dedicated memory region that
sequentially arranges needed KV vectors for attention computation.  The content of
the execution buffer is copied from the steady zone, block cache, and directly
from the CPU memory in case of a cache miss (details in
Section~\ref{subsec:wave-buffer}).  Second, it acts as a CPU-resident buffer manager that manages
the block cache and data movement between GPU and CPU memory.

During decoding, \newsysname{} computes the attention for each head in
parallel, following the steps in Figure~\ref{fig:arch}:
\circleb{1}: The centroids are sorted according to the similarity to the query
vector, determining a subset of more critical clusters to retrieve for precise
attention computation, and the clusters to perform estimation.
\circleb{2}: The GPU performs attention estimation (2-G), and a request is sent
to the buffer manager to retrieve the needed clusters (2-C).
\circleb{3}: The buffer manager ensures that the KV blocks are ready in the
execution buffer through parallel data copying.  
\circleb{4}: The GPU computes the precise attention using the KV vectors in the execution buffer, while the attention estimation (2-G) result is merged.

\begin{figure}[t!]
    \centering
    \includegraphics[width=0.96\linewidth]{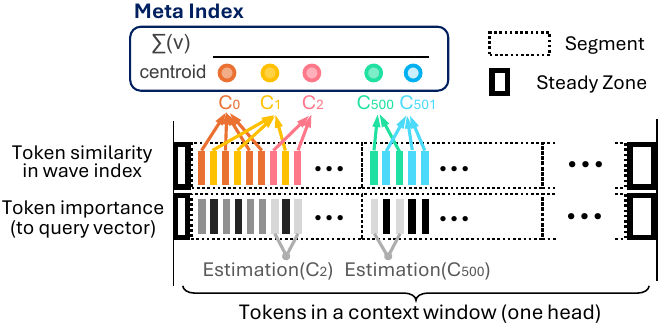}
    \caption{Attention-aware design of wave index.
    The wave index partitions the context into segments (dotted boxes), where each vertical bar denotes a token. In the first row, tokens with the same color form a cluster, and similar colors (e.g., yellow and orange) indicate key vector similarity. In the second row, darker bars indicate higher importance to the query. For example, a purple query may be close to a red vector (1st segment) and a blue vector (2nd segment).}
    \label{fig:waveindex}
\end{figure}

\subsection{Wave Index}
\label{subsec:wave-index}
Wave index employs a cluster-based vector index and designs for the
attention mechanism to ensure accurate and efficient computation.

\minisec{Index Organization}
Wave index partitions the KV vectors in the context into clusters using spherical
$k$-means~\cite{hornik2012spherical}, based
on the similarity between key vectors. We choose spherical $k$-means because it is widely used for inner-product-based clustering~\cite{douze2024faiss}, aligning with the similarity metric between query and key vectors in attention mechanisms.
Each cluster is represented by its centroid, computed as the mean of its key vectors.
We design the wave index based on a cluster-based vector index for two
reasons. First, centroids serve as compact representatives and can reside in fast GPU memory, aligning well with CPU-GPU asymmetry.
Second, clustering improves access locality by reducing random
accesses over PCIe. 
As illustrated in Figure~\ref{fig:waveindex}, although several similar tokens (i.e.,
2nd, 4th, and 8th) are irregularly distributed 
in the context, wave index groups them into the same cluster (i.e., yellow).

When serving a query vector, wave index sorts the centroids based on the inner
product between the query vector and the centroids, as the inner product determines
the attention score. Classic vector indexes have addressed
the complexity of high-dimensional vector spaces to effectively perform
clustering and compute centroids~\cite{hornik2012spherical,douze2024faiss}. 
However, the goal of classic ANNS is to
retrieve the top-$k$ vectors by selecting candidates from clusters and performing
pairwise comparisons to the vectors for further ranking, 
which is costly on GPU. In contrast, wave index
retrieves all vectors from the critical clusters and incorporates them directly
into the attention computation, avoiding the selection cost.
Since the clusters are highly coherent, this design also improves the
attention accuracy by potentially including more tokens that are also important.

To accelerate index construction, we introduce a technique called
{\textit{segmented clustering}}, where the input sequence is divided into
segments, and clustering is performed within each segment, instead of
clustering over the entire input sequence as in classic vector index.
For example, in Figure~\ref{fig:waveindex}, vectors in the first
segment (i.e., red, yellow, and orange ones) are clustered  
independently of vectors in the second segment.
We describe its rationale and details later.

For physical placement, the meta index, which requires only a small memory
footprint, resides in GPU memory. The meta index contains a table of centroids, the sum of the value vectors and the cluster
size, as shown in Figure~\ref{fig:arch}. The large volume of key and value
vectors resides in CPU memory, organized into contiguous KV blocks. This
block-based organization facilitates efficient data movement between GPU and CPU
memory.

\minisec{Tripartite Attention Approximation}
We introduce {\textit{tripartite attention approximation}} to ensure the
accuracy of attention computation with reduced computation cost by leveraging both the properties of the
attention mechanism and the advantage of vector retrieval. The final KV vectors
used for attention computation consist of three parts: {\textit{steady zone}},
{\textit{retrieval zone}}, and {\textit{estimation zone}}, and we denote the
attention output as $o^0$, $o^1$, and $o^2$ respectively.

\minisubsec{Steady, Retrieval, and Estimation Zones}
Recent studies have revealed that a range of tokens located at the beginning and end
of the context are consistently important~\cite{streamingllm}.
They are categorized as steady zone in wave index, as shown
in Figure~\ref{fig:waveindex}.  Therefore, tokens in the steady zone are
directly included based on their positions, rather than being organized into the
vector index.  
The KV vectors retrieved by the centroids and clusters are
categorized as retrieval zone; these tokens are directly used for attention computation. 

Due to the high variety of sparsity ratios across model layers, attention heads,
queries, and task types, determining the optimal number of clusters for the retrieval zone is challenging, if not unattainable.
To compensate for the potential accuracy loss of non-retrieved clusters, we introduce an accuracy-bounded attention estimation mechanism.
Tokens in the non-retrieved clusters are categorized as the estimation zone.
This approach effectively handles sparsity variation without increasing data transfer over PCIe.

\begin{figure}[t!]
    \centering
    \includegraphics[width=\columnwidth]{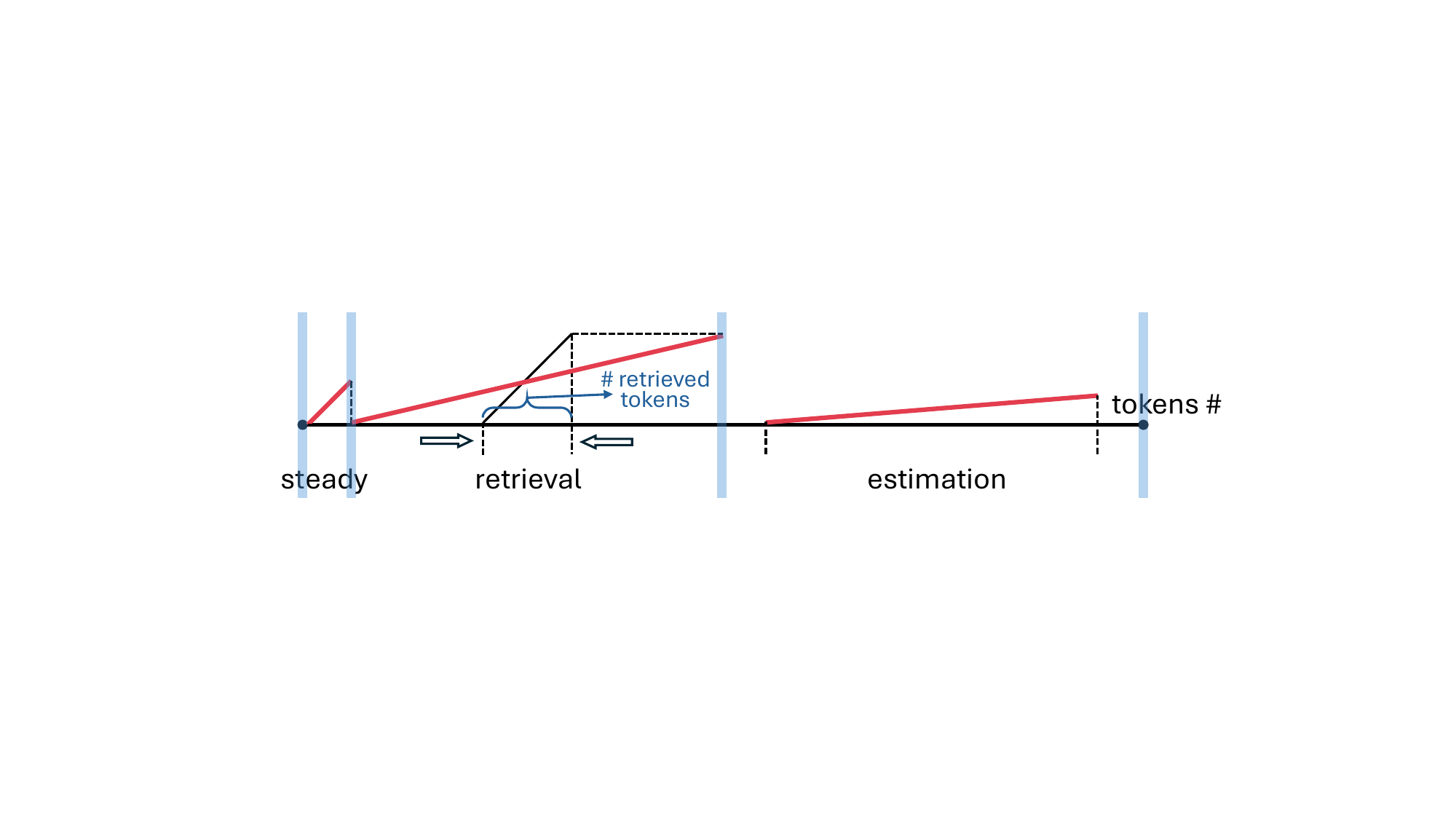}
    \caption{
    Illustration of computation cost for three zones.
    The x-axis denotes the token count, while the y-axis denotes the computation cost.
    In the steady zone, attention is computed precisely, so the computation cost
    is linear in the number of tokens (red line's slope).  
    In the retrieval zone, we calculate
    attention for the retrieved tokens (whose number is significantly reduced
    due to sparsity). The computation cost is linear in the number of retrieved
    tokens (black line), while the overall cost is amortized over the total number of
    vectors (red line).  In the estimation zone, the cluster token's 
    attention is estimated based on centroids, which incurs a much lower
    computation cost (red line).
    }
    \label{fig:3zone-model}
\end{figure}

Combining the three zones, we have $p$ KV vectors in the steady zone,
and $m$ clusters in the retrieval and estimation zones,
with $r$ clusters assigned to the retrieval zone and the remaining assigned to the estimation zone.
The $m-r$ clusters in the estimation zone enhance the
robustness of the chosen $r$ clusters for retrieval in the face of
varying sparsity ratios, thus improving attention accuracy.
Intuitively, our tripartite attention approximation aligns with the inherent sparsity of attention. 
From $o^0$ to $o^2$, the attention weight computation becomes progressively lighter and more approximate, while the resulting values decrease in magnitude, as illustrated in Figure~\ref{fig:3zone-model}. 

\begin{figure}[t!]
    \centering
    \includegraphics[width=.97\columnwidth]{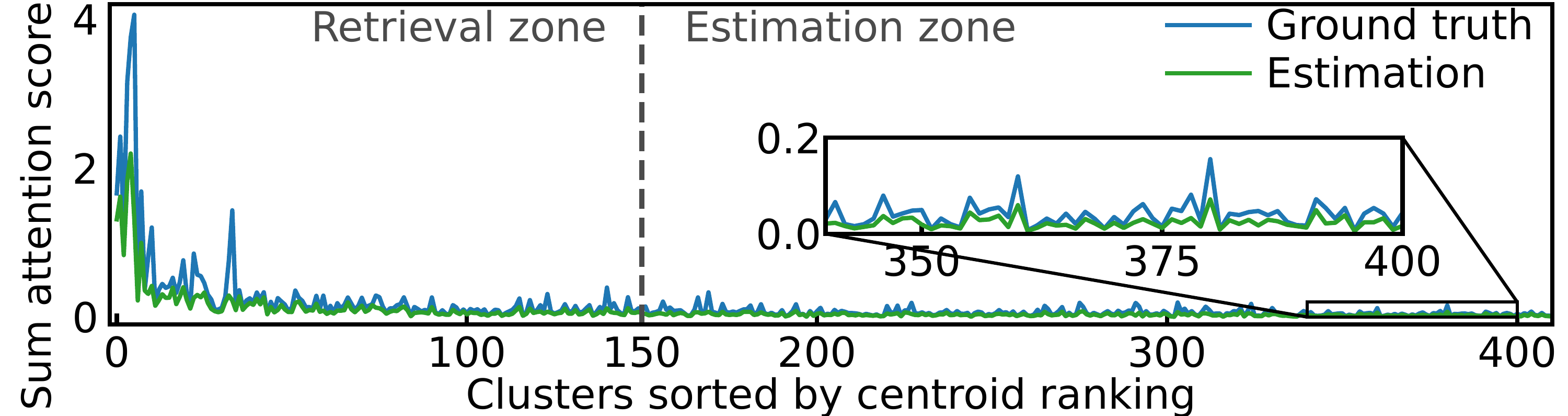}
    \caption{
    Centroid representativeness and estimation accuracy (Llama3-8B-1048K, layer 30, head 24, \textit{fwe}, 128K context).
    The x-axis is the ranking of centroids during index traversal. The blue line indicates that top-ranked centroids have higher cumulative attention scores, showing the effectiveness of clustering.
    Attention estimation (green) follows the same trend of ground truth with a tight bound.
    }
    \label{fig:estimation}
\end{figure}

The clusters exhibit high coherence and strong alignment with attention scores,
along with highly representative centroids (Figure~\ref{fig:estimation}),
enabling a robust distinction between the retrieval and the estimation zone.
Precise attention computation is critical for top-ranked centroids, which have a greater impact on the output. 
In contrast, the estimation zone targets clusters with lower attention scores, where lightweight approximations can be applied to significantly accelerate computation.
This design ensures the efficient retrieval of important tokens while maintaining accuracy by bounding the influence of tokens in the estimation zone.

Sensitivity analysis in Section~\ref{subsec:micro} shows that using a small fixed steady zone, a small retrieval zone and a larger estimation zone consistently delivers both high throughput and high accuracy across different tasks and models.

\minisubsec{Accuracy-Bounded Attention Estimation}
\label{subsec:design_est}
The key idea for accuracy bounded attention estimation is to
use the centroids to estimate the attention weights for all KV vectors
within a cluster, thereby avoiding the computation of
attention for each individual KV vector, striking a balance between
accuracy and efficiency.

Consider an attention head with $p$ KV vectors in the steady
zone and $m$ centroids, denoted as $ \mathbf{C} \in \mathbb{R}^{m \times d} $, with
corresponding cluster sizes $ \mathbf{s} \in \mathbb{R}^{m \times 1} $. The attention
weight for the KV vectors in the $i$-th cluster is estimated by:
\begin{equation}
    \tilde{a}_{i} = \dfrac{
        e^{ \mathbf{q} \cdot \mathbf{C}^T_i / \sqrt{d} }
    }{
        \sum_{j=1..p}{ e^{ \mathbf{q} \cdot \mathbf{K}^T_j / \sqrt{d} } } + 
        \sum_{j=1..m}{\left(s_j \cdot e^{ \mathbf{q} \cdot \mathbf{C}^T_j / \sqrt{d} }\right)}
    },\ i = 1..m
    \label{eq:estimate0}
\end{equation}

The accuracy of this estimation is assured because the centroid estimation acts
as a lower bound for the sum of exponential inner product values within a
cluster. For the $i$-th cluster with $s_i$ key vectors, the centroid is the average of all key vectors in the cluster. By applying Jensen's
inequality~\cite{jensen1906fonctions}, 
we derive the following bound:
\begin{equation}
    \mathbf{C}_{i} = \dfrac{
        \sum_{j=1..s_i}{\mathbf{K}_j} 
    }{
       s_i
    },
    \quad
   e^{ \mathbf{q} \cdot \mathbf{C}^T_i / \sqrt{d} } \leq 
   \dfrac{ \sum_{j=1..s_i}{ e^{ \mathbf{q} \cdot \mathbf{K}^T_j / \sqrt{d}} } }{s_i} 
    \label{eq:estimate1}
\end{equation}

Figure~\ref{fig:estimation} illustrates the accuracy bound of the estimation. 
Notably, while clusters are ranked by centroids ($\mathbf{q} \cdot \mathbf{C_i}^T$), 
the estimation is not monotonically decreasing because the sum of attention scores 
within a cluster is influenced by the non-uniform cluster sizes ($s_i$).

To further reduce the overhead of accessing corresponding value vectors, 
we sum the value vectors of each cluster during the index construction and store
cluster size and summed value vectors ($VS \in \mathbb{R}^{m \times d}$)  in the
meta index.  This approach avoids individual value access during inference and
is based on the following equality for the partial attention output
$\tilde{o}_i$ derived from a cluster $C_i$ and the attention output $o_2$ for all
non-retrieved clusters:
\begin{equation}
    \mathbf{\tilde{o}_i} = \sum_{j=1}^{s_i} (\tilde{a}_{i}\cdot \mathbf{V}_j) = \tilde{a}_{i} \sum_{j=1}^{s_i}\mathbf{V}_j = \tilde{a}_{i} \cdot VS_i,
    \quad
    \mathbf{o_2} = \sum_{i=1}^{m-r}\mathbf{\tilde{o}_i}
    \label{eq:attn-detail}
\end{equation}

As such, the time complexity of attention estimation for non-retrieved clusters
is $O(m-r)$, which is substantially lower than accessing individual KV vectors,
ensuring the efficiency of our approach, particularly for longer contexts.

\minisec{Lightweight Index Construction and Updates}
\label{subsec:index_const}
Index construction occurs during the prefilling phase, where the full KV
vectors are offloaded to CPU memory, and clustering is performed on the
GPU to produce the meta index. The main challenges in index construction are
reducing computation overhead to prevent slowing down the prefilling.
While cluster-based vector index incurs lower index construction
cost among vector index approaches, performing $k$-means across the
entire input sequence is still expensive.

To save construction cost, wave index introduces \textit{segmented clustering}, performing $k$-means
clustering within each segment independently, reducing the time complexity of clustering
by a factor of $s$, where $s$ is the number of segments. This design is based on the
observation that key vectors exhibit coarse-grained spatial locality within the input
sequence\footnote{We attribute the spatial locality to
RoPE~\cite{su2024roformer}, which encodes positional information in key vectors.
This effect is confirmed by observing significantly less locality when
performing $k$-means clustering on Pre-RoPE versus Post-RoPE key vectors.}.
Key vectors within a broader region (i.e., a single segment)
tend to be similar to each other but dissimilar to those in other segments,
which negates the need to include distant keys during clustering.
As shown in Figure~\ref{fig:waveindex}, tokens from the three
clusters (i.e., yellow, red, and orange) are relatively similar.
Though clustering results may vary across different contexts or
attention heads, tokens in the first segment are unlikely to be similar to those
in the second segment (e.g., green and blue tokens).
Section~\ref{sec:segment_exp} presents the analysis of segmented clustering,
where an 8K segment size achieves index quality comparable to global $k$-means,
while significantly reducing the build cost.

Such coarse-grained spatial locality does not contradict the fact 
that important tokens (i.e., clusters) are usually scattered throughout the context (Figure~\ref{fig:waveindex}).
This is due to the fact that similarity-based clustering is 
conducted in a high-dimensional vector space and the query vector in attention 
behaves as a special type of ``out-of-distribution query vector.''~\cite{liu2024retrievalattention}
In this context, the key vectors similar to the query vector can possibly 
be distant from each other in the vector space~\cite{roar}.
We address this with a classic centering technique~\cite{all-top} 
(inspired by MagicPIG~\cite{magicpig})
to ensure the clustering effectively captures the attention importance.

The spatial locality also helps in managing newly generated tokens during decoding. 
New tokens are appended to the steady zone, and $k$-means clustering is applied
once the local window reaches the size of a segment. 
This design appends new clusters into the index, avoiding full re-clustering 
and significantly reducing index update overhead.
We empirically set the update segment size to 1K tokens through extensive experiments, 
a choice that well balances the inference accuracy and efficiency. 
Since clustering is performed once every 1024 tokens, 
the clustering cost is effectively amortized among generated tokens. 
Our evaluation shows that index update contributes only a 0.2\% latency overhead during decoding. 
A larger segment size increases GPU memory consumption
for buffering tokens, reducing batch size and limiting decoding throughput.

\subsection{Wave Buffer}
\label{subsec:wave-buffer}
With the wave index, \newsysname{} significantly reduces the number of KV
vectors required per query, lowering PCIe traffic to less than $2\%$
of that in full attention. However, the inference throughput could still suffer
due to the limited PCIe bandwidth for long-context inference. Drawing on the
classic wisdom of computer systems, the performance and capacity disparity
between GPU memory and CPU memory, akin to the von Neumann bottleneck~\cite{von1993first},
necessitates the use of buffer caches in the fast, low-capacity GPU memory. 
Thanks to the high-quality retrieval provided by the wave index, adjacent
decoding steps for neighboring query vectors show strong temporal locality, as
only important tokens are loaded into GPU memory. Thus, caching is highly
beneficial.

The organization of the wave index, combined with the hardware characteristics,
poses unique challenges in managing the memory used in \newsysname{} to
achieve better throughput. Specifically, what should be cached, how to manage
data movement and the control plane of caching, and finally, how can all these
movements be effectively parallelized? We address these challenges with the wave
buffer, a comprehensive memory management component that aligns well with the
attention mechanism.

\begin{figure}[t!]
    \centering
    \includegraphics[width=0.95\columnwidth]{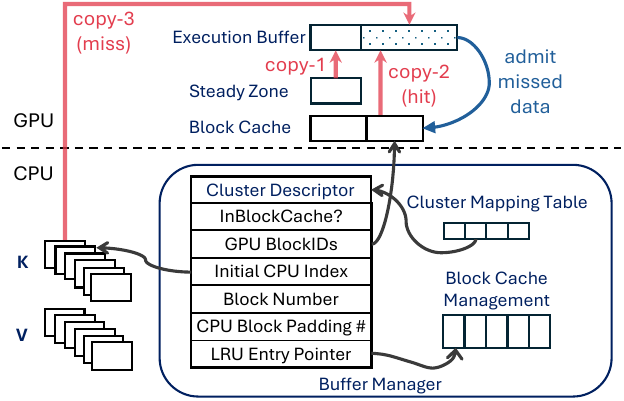}
    \caption{Design of wave buffer.
    Black arrows denote pointer relationships between data structures. Red arrows indicate the three possible sources for assembling the execution buffer. Missed data is admitted into the block cache by copying from the execution buffer (blue arrow).}
    \label{fig:cache}
\end{figure}

\minisec{Buffer Organization}
Within GPU, wave buffer contains a block cache for the KV vectors in the
retrieval zone, a small buffer for the steady zone, and an attention-layer-share execution buffer. 
The control plane of the wave buffer resides in CPU memory, with its data structures managed by the buffer manager running on a CPU thread pool.
Figure~\ref{fig:cache} shows the
organization of wave buffer.

\minisec{KV Block Cache}
\label{subsubsec:blockcache}
We observe a strong temporal locality of important tokens: neighboring decoding
steps (i.e., query vectors) have a high overlap of critical tokens due
to topic continuity and syntactic proximity. 
With a GPU cache size equal to 5\% of the total KV vectors, the average cache hit ratio across tasks ranges from 0.79 to 0.94, indicating high cache ability of important KV vectors.
This observation is also supported by recent works~\cite{tokenselect, xu2024recycled}. 

However, exploiting this temporal locality in the KV Block Cache
is challenging for two reasons. 
First, a mismatch exists between the cluster operations and memory management units.
Clusters are suitable as the \textit{logical unit} of cache access
to align with the wave index's cluster-based operations,
while fixed-size KV blocks are preferred as the \textit{physical unit} of storage for simplicity in memory management.
Since a single cluster can span multiple blocks, 
a semantic gap arises between these two units.
To bridge it, we introduce a cluster mapping table as a layer of indirection,
implemented as an array indexed by cluster ID for fast access.
Each entry is a cluster descriptor (Figure~\ref{fig:cache}), 
mapping the logical cluster to its physical block addresses in GPU cache and CPU memory. 
This design enables efficient translation
from cluster-level retrieval to block-level data movement
while preserving management simplicity.

Second, cache management overhead must remain far below the per-layer decoding latency to maintain the inference efficiency. 
Specifically, at a 128K context length, one layer of Llama3-8B executes in 720~$\mu$s on the GPU. 
In contrast, integrating a standard LRU cache implementation with PyTorch-based data transfer introduces roughly 1.5~ms of additional overhead per layer, 
which offsets any potential benefit from reduced PCIe traffic.
To meet this stringent latency requirement, wave buffer introduces two optimizations.
First, we implement dedicated GPU kernels for efficient GPU–CPU data transfers (detailed in Section~\ref{subsec:impl}), reducing the copy time from 1.3~ms to 100~$\mu$s under cache misses.
Second, we move the cache replacement logic to the CPU, decoupling cache access 
on the critical path from expensive cache updates.
This is facilitated by the hybrid
GPU-CPU structure--after the lookup, the KV vectors must be immediately
ready for the GPU in the execution buffer, whereas the cache update can be
performed asynchronously by the CPU, in parallel with the data 
copy and attention computation. 
Compared to GPU-native cache management~\cite{wang2022merlin}, 
our CPU-managed, asynchronously updated cache achieves low latency requirement, 
simplicity, and better extensibility for various caching policies.

\minisec{Synchronous Cache Access} 
The mapping table associates each cluster ID with its block IDs 
in either GPU memory (cached) or CPU memory (missed). 
Upon obtaining the cluster IDs in the retrieval zone for a query, 
the wave index sends them to the CPU thread running the buffer manager, 
which looks up the mapping table to determine the clusters' 
locations and issues the memory copy accordingly. 
Table lookup is accelerated using CPU multi-threading, 
without synchronization overhead, as this process is read-only.
As shown in Figure~\ref{fig:arch}, we parallelize the mapping table lookup with
attention estimation on the GPU to reduce inference latency.

\minisec{Asynchronous Cache Update}
Following the mapping table lookup, the wave buffer immediately submits a cache 
update request to the CPU thread pool for asynchronous execution. As a result,
cache updates are decoupled from cache access, allowing the wave buffer to issue
memory copy operations via GPU threads while concurrently making cache
replacement decisions and metadata updates (e.g., LRU lists) on the CPU. The
replacement decision is then enacted by issuing a request to the GPU to schedule
a copy from the execution buffer to the block cache, thereby admitting the data
into GPU memory and completing the cache update.

Decoupling cache access and update allows CPU latency to overlap with GPU execution, accelerating the overall process.

\minisec{Assemble the Execution Buffer}
The execution buffer contains the final key-value vectors in the steady zone and
the retrieval zone, arranged in contiguous memory. The sequential nature of the
execution buffer makes them directly usable by FlashAttention~\cite{dao2022flashattention} to 
compute the partial attention output. This output is combined
with the estimated attention to produce the final attention
output.

Three types of data copy may be needed to assemble the execution buffer, as shown in Figure~\ref{fig:cache}:
1) from the steady zone (GPU to GPU),
2) from the KV block cache (GPU to GPU), and
3) from KV blocks (CPU to GPU).
We have developed a highly-optimized copy operator 
to accelerate the copy process (details in Section~\ref{subsec:impl}).

\subsection{Constrain Prefilling Latency}
Because prefilling is compute-intensive, special care is required to perform
all buffer-related computation (e.g., construct the mapping table) alongside
the lightweight index construction. We carefully constrain prefilling latency
by fully utilizing GPU-CPU parallelism.

After obtaining the Q, K, and V during prefilling, \newsysname{} asynchronously
offloads the KV vectors to CPU memory while the GPU performs segmented
clustering. The clustering results are then sent to the CPU, allowing the wave
buffer to asynchronously construct its data structures, including the
mapping table and metadata for cache replacement, in parallel with
attention calculation. With this parallelism, only segmented clustering remains
on the critical path of the prefilling phase, and its latency is negligible
(less than 5\%, as shown in Section~\ref{subsec:prefill_latency}).

\subsection{Scale to Multiple GPUs}
Serving inference for larger LLMs typically adopts
model partitioning across multiple GPUs and model-parallel
execution~\cite{alpaserve,alpa}. \newsysname{} extends naturally to this setup
as the wave index and wave buffer are modular components co-located with
each attention head, requiring no coordination across layers or heads. In
multi-GPU setting, \newsysname{} integrates with partitioned
layers or heads and manages the corresponding index and buffer on each GPU.
Section~\ref{subsec:efficiency} demonstrates its effectiveness.

\subsection{Implementation}
\label{subsec:impl}
We implement a custom Triton~\cite{tillet2019triton} kernel for segmented clustering in the wave index. The kernel accepts KV vectors and executes $k$-means in parallel across attention heads and segments. It then computes the value sums and stores the results in the meta index.

GPU-side memory copies for accessing and updating the KV block cache and execution buffer are complicated, as source and destination addresses are often non-contiguous and may involve both CPU–GPU and GPU–GPU transfers. We implement CUDA kernels (\textasciitilde 1,000 LoC) for efficient execution buffer transfers and block cache updates, skipping over fragmented regions within blocks to mitigate fragmentation.  Since attention heads may exhibit varying copy patterns due to differing cache hit ratios, number of threads are dynamically changed to maintain performance.

Attention in the estimation zone differs from standard attention, as cluster sizes are required to compute attention weights. We modify FlashAttention~\cite{dao2022flashattention} kernel to support weighted attention, which also efficiently merges attention outputs from three zones.

\section{Evaluation}
\label{sec:eval}

We thoroughly evaluate \newsysname{}, comparing it with full attention and state-of-the-art sparsity-based KV cache systems across various models and tasks. The results demonstrate that:
\begin{itemize}
    \item \newsysname{} outperforms sparsity-based baselines in model accuracy, and is the only system that matches full attention accuracy.
    \item \newsysname{} achieves high throughput while scaling the context lengths and batch size.
    \item The techniques incorporated into the wave buffer, such as GPU caching and asynchronous cache updates, effectively minimize data transfer over PCIe and boost throughput.
\end{itemize}

\subsection{Experimental Setup}
\label{subsec:setup}
We run experiments on a VM server with NVIDIA A100 GPUs (80GB memory) and an
AMD EPYC 7V12 CPU (1.7TB memory). It has 4 NUMA nodes (12 cores each).
The GPU and CPU are linked via PCIe 4.0 ($\times$16), with a 
unidirectional bandwidth of 32GB/s. The server runs Ubuntu 22.04 
with CUDA 12.4 and PyTorch 2.5. 

\minisec{Models}
We evaluate \newsysname{} on four open-source LLMs: 
Llama3.1-8B~\cite{llama3.1}, 
Qwen2.5-7B~\cite{qwen2.5}, Llama3-8B-1048K~\cite{llama3-8B-1048k} and Qwen2.5-72B~\cite{qwen2.5-72b}. 
Llama3.1-8B, Qwen2.5-7B and Qwen2.5-72B support up to 128K tokens,
while Llama3-8B-1048K supports a 1048K context window.
Unless otherwise specified, all models are evaluated on a single A100 GPU except that Qwen2.5-72B is evaluated 
on eight A100 GPUs by evenly partitioning the model layers across GPUs. 
We include two advanced reasoning models DeepSeek-R1-Distill-Llama-8B~\cite{deepseek-llama} 
and DeepSeek-R1-Distill-Qwen-7B~\cite{deepseek-qwen}, to evaluate \newsysname{} on long-generation reasoning scenarios.

\minisec{Baselines}
We compare with a highly-optimized full attention baseline that uses 
FlashInfer~\cite{ye2025flashinfer} in both accuracy and efficiency. 
vLLM~\cite{kwon2023efficient}, which internally relies on full attention, 
is included for end-to-end latency evaluation in Section~\ref{subsec:e2e}. 
We also compare \newsysname{} with 
four state-of-the-art dynamic sparsity-based KV cache systems: Quest~\cite{quest}, MagicPIG~\cite{magicpig}, 
InfiniGen~\cite{lee2024InfiniGen}, and PQCache~\cite{zhang2024pqcache}. 
Quest is GPU-only and employs chunk-based retrieval based on representative vectors, while 
other three systems are GPU-CPU frameworks which offload KV cache to CPU memory.
MagicPIG employs LSH to sample relevant tokens and relies on CPU for most computation to reduce PCIe transfer.  
In contrast, InfiniGen speculatively prefetches important tokens to GPU based on similarity between adjacent layers.
PQCache, a recent system that applies ANNS to attention, leverages product quantization (PQ) 
to identify important tokens and load them to GPU for computation. 
We use their open-sourced codes~\cite{quest-code, magicpig-code, infinigen-code, pqcache-code} 
for evaluation. 

\minisec{Parameters} 
Each sparsity-based baseline uses the parameters from its original paper\footnote{Quest uses a chunk size of 16 and applies full attention to the first two model layers; 
MagicPIG employs 10 hash functions and 150 hash tables and applies full attention in selected layers; 
InfiniGen uses 32 partial channels; 
PQCache uses 2 partitions and 6-bit PQ codes for context $\leq$ 16K and switches to (4, 8) for context $>$ 16K.}.
\newsysname{} uses one centroid per 16 tokens on average,
with segments of 8K tokens and 10 iterations for $k$-means to balance index quality and build time. 
The retrieval budget is set as $1.8\%$\footnote{For \newsysname{}, in the 128K context with 8192 clusters, we set 150 clusters in the retrieval zone, so 
the retrieval budget is $\frac{150}{8192}\approx 1.8\%$.} across different 
sparsity-based systems and tasks, aligning with the budget used in~\cite{magicpig} for fair comparison.
For \newsysname{}, the steady zone contains 4+64 tokens for initial and local windows while the estimation zone is
set as 23.2\% of total clusters so that three zones cover $\sim$25\% of the context.
We study the impact of different zone sizes in Section~\ref{subsec:micro}.
For wave buffer, 
we explored several cache policies and selected LRU as default 
due to its best performance. The GPU cache size is empirically set as 
5\% of all KV vectors and the block size is set as 2KB.
We use 24 logical threads in one NUMA node.  
MagicPIG uses all cores from two NUMA nodes, following 
its original setting~\cite{magicpig-code}. 

\minisec{Benchmarks}
We utilize two representative long-context benchmarks for end-to-end accuracy evaluation.
(1) RULER~\cite{hsieh2024ruler}: a comprehensive benchmark covering 13 different tasks such as 
aggregation and question answering. Each task contains 200 examples, under context lengths 
from 8K to 128K. 
(2) Needle-in-a-haystack (NIAH)~\cite{needle}: it challenges LLMs to retrieve 
critical information (``needle'') from a context (``haystack'').
We use NIAH to stress model's accuracy up to 1 million tokens. 
End-to-end task accuracy is used to show the inference quality of \newsysname and baselines.

We also utilize two widely used reasoning benchmarks, 
AIME24 \cite{aime24} and GPQA~\cite{gpqa}, 
to evaluate long-generation ability, which feature 
inputs with a few hundred tokens and up to 32K output tokens. 
For GPQA, we uniformly sample 50 examples for evaluation.

We evaluate \newsysname{} and other baselines' inference efficiency with different 
batch sizes and context lengths. By default, we use Llama3-8B-1048K and NIAH to 
test \newsysname{} up to 1 million tokens. Additional models and tasks and end-to-end evaluations are 
also covered in Section~\ref{subsec:efficiency}.

\begin{figure}[t!]
    \centering
    \includegraphics[width=\columnwidth]{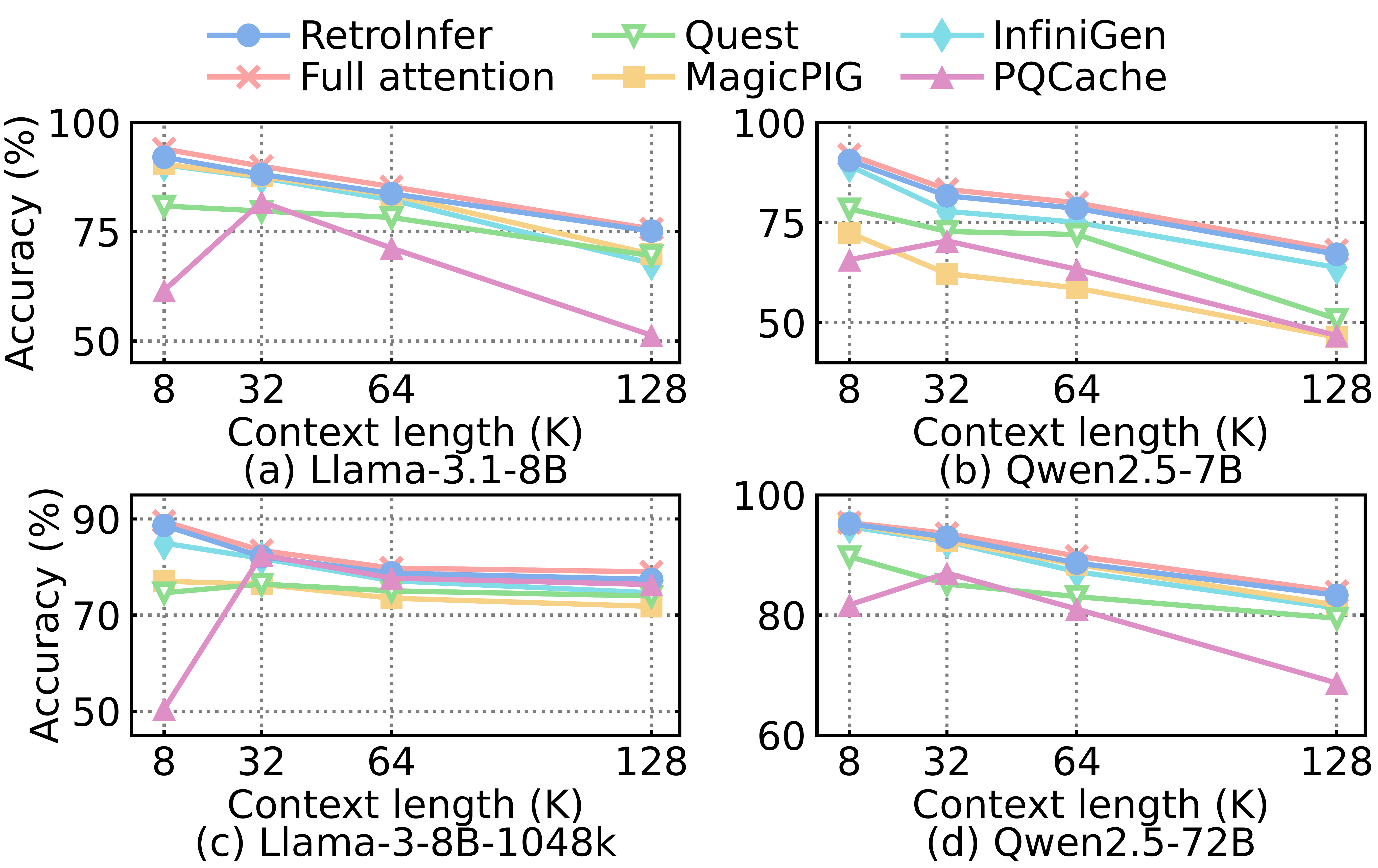}
    \caption{RULER accuracy under different context lengths and models. \newsysname is the only solution that matches full attention accuracy across different context lengths.}
    \label{fig:ruler_acc}
\end{figure}

\subsection{Inference Accuracy}
\label{subsec:accuracy}
We first analyze the end-to-end accuracy on different benchmarks. 

\minisec{RULER and NIAH}
Figure~\ref{fig:ruler_acc} shows the average task accuracy of different 
systems under different context lengths on RULER.
For all context lengths and models, \newsysname{} consistently matches the 
accuracy of full attention and outperforms all sparsity-based baselines.
At 128K context, \newsysname{} has only a drop of 
0.66\%\slash 1.50\%\slash 1.97\%\slash 0.73\%
compared to full attention on Llama3.1-8B\slash Qwen2.5-7B\slash Llama3-8B-1048K\slash Qwen2.5-72B, respectively.
Compared to the sparsity-based systems, \newsysname{} achieves accuracy improvements 
ranging from 1.40\% to 46.47\%.
These results indicate that \newsysname{} is able to effectively retrieve important 
tokens using wave index and cover the varying sparsity by estimation robustly.

\begin{figure}[t!]
    \centering
    \begin{minipage}{0.37\columnwidth}
        \centering
        \includegraphics[width=.85\textwidth]{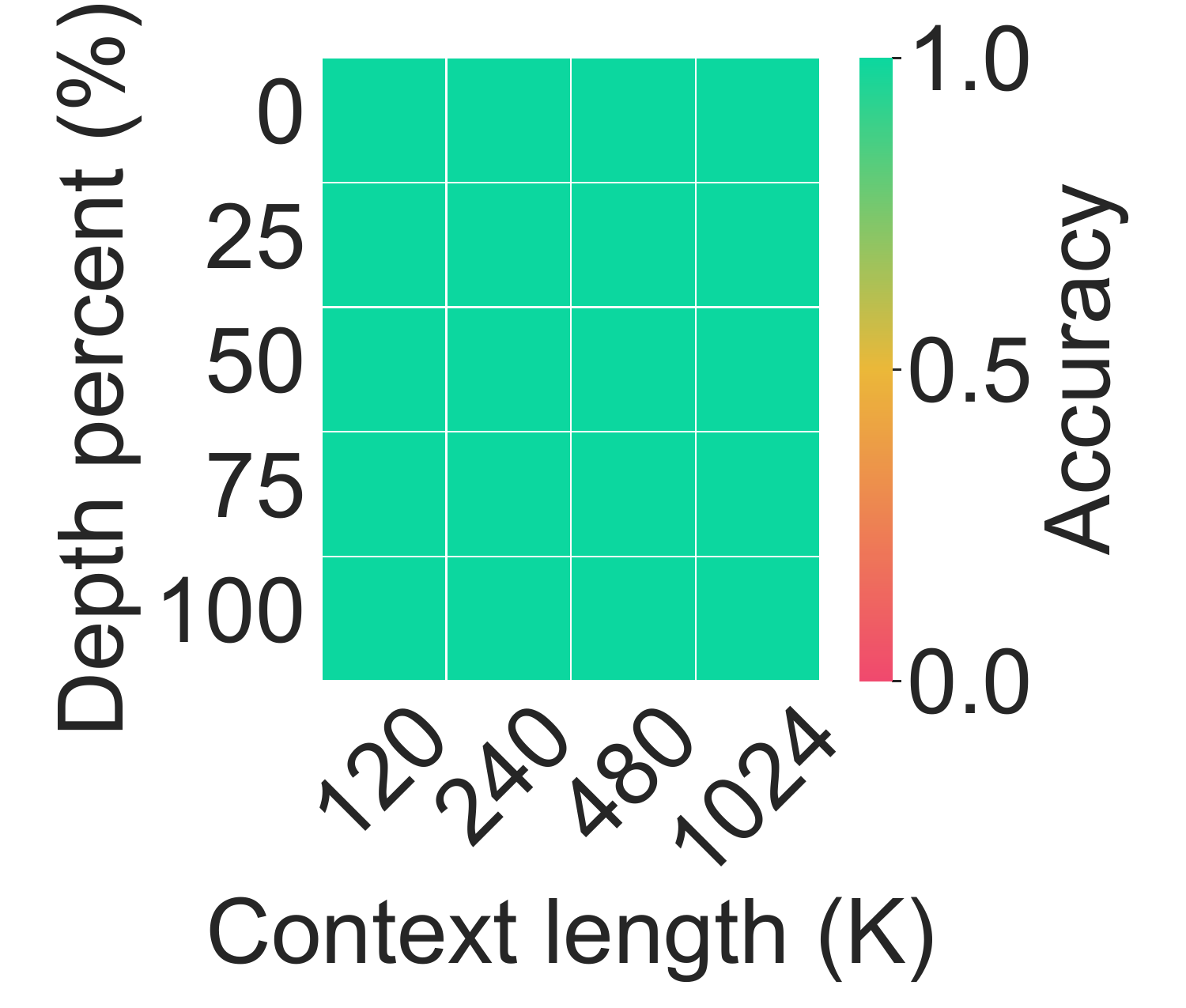}
        \caption{Needle-in-a-haystack results on Llama3-8B-1048K. 
        }
    \label{fig:niah}
    \end{minipage}
    \hfill
    \begin{minipage}{0.56\columnwidth}
        \centering
        \includegraphics[width=\textwidth]{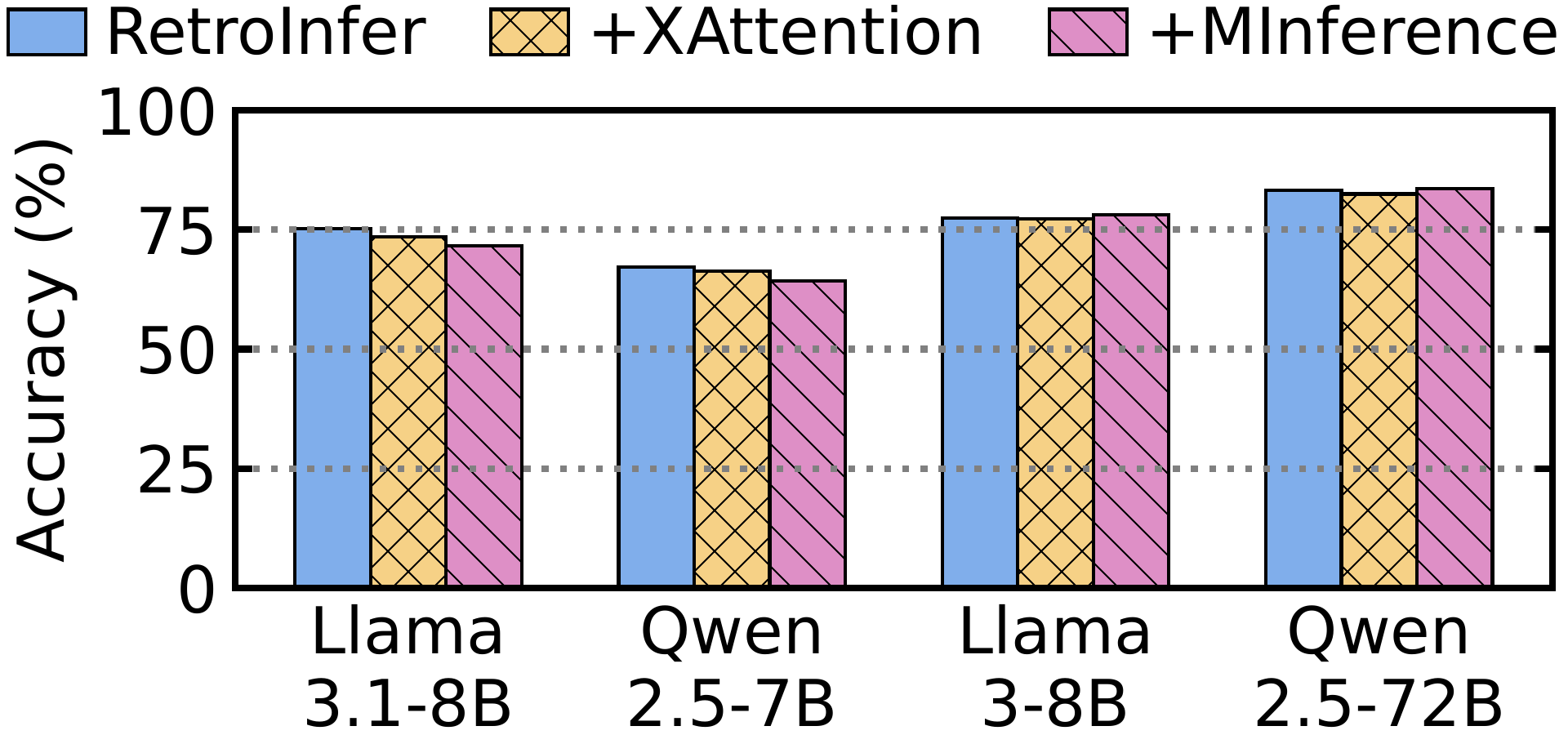}
        \caption{RULER accuracy (128K) of \newsysname{} combined with XAttention and MInference. 
        }
    \label{fig:x_attn}
    \end{minipage}
\end{figure}

Figure~\ref{fig:niah} further shows that \newsysname{} achieves 100\% accuracy
with context lengths up to 1 million tokens.
This demonstrates our system's capability to support million-token with high accuracy.

\begin{table}[t!]
    \centering
    \caption{Reasoning tasks accuracy (pass@8~\cite{nilearning}). \newsysname achieves comparable accuracy to full attention.}
    \label{tab:long_gen}
    \resizebox{\linewidth}{!}{
    \begin{tabular}{l|cc|l|cc}
    \toprule
    \textbf{Systems} & AIME24 & GPQA & \textbf{Systems} & AIME24 & GPQA \\
    
    \midrule
    \textit{DeepSeek-Llama} & \textit{73.30} & \textit{68.00} & \textit{DeepSeek-Qwen} & \textit{76.70} & \textit{72.00} \\
    Quest & 46.70 & 50.00 & Quest & 46.70 & 62.00 \\
    InfiniGen & 36.70 & 62.00 & InfiniGen & 60.00 & 66.00 \\
    PQCache & 30.00 & 58.00 & PQCache & 13.33 & 46.00 \\
    \rowcolor{blue!5}
    \newsysname{} & \textbf{76.70} & \textbf{68.00} & \newsysname{} & \textbf{73.30} & \textbf{68.00} \\
    
    \bottomrule
    \end{tabular}
    }
\end{table}

\minisec{Reasoning Tasks}
\label{subsec:long_gen}
We evaluate \newsysname{} on two reasoning benchmarks~\cite{aime24, gpqa} using reasoning models~\cite{deepseek-llama, deepseek-qwen}.
The evaluation follows the standard settings~\cite{guo2025deepseek} and 
sets the maximum generation length as 32K.
Since the input length is relatively short, we do not construct 
indexes during the prefilling stage. 
As the generation progresses, once the combined length of prompt and 
generated tokens exceeds 1K, the index is initialized and 
then updated incrementally with an update segment size of 1K tokens.
We exclude MagicPIG from the comparison, as it does not support index updates for long-generation tasks.
Table~\ref{tab:long_gen} shows that \newsysname matches the accuracy of full attention and achieves higher accuracy compared to the baselines,
demonstrating \newsysname{}'s strong capability for long generation.
Notably, \newsysname{} achieves higher accuracy than full attention in some tasks.
This is because the sparsity mechanism helps filter out less informative tokens generated by
the reasoning process~\cite{yang2025understanding, chen2024not}, 
allowing the model to attend to more relevant context and improve generation quality.

\minisec{Compatibility with Sparse Prefilling}
\newsysname{} and other baselines focus on optimizing the decoding phase. 
Now we study \newsysname{}'s compatibility with the state-of-the-art sparse 
prefilling methods XAttention~\cite{xuxattention} and MInference~\cite{jiang2024minference}, 
which aim at reducing prefilling latency. 
We evaluate \newsysname{} on RULER with and without them, as shown 
in Figure~\ref{fig:x_attn}.
The accuracy \newsysname{} combined with XAttention or MInference only drops 
by 1.52\% on average. This demonstrates that \newsysname{} is compatible with 
prefilling acceleration techniques.

\subsection{Inference Efficiency}
\label{subsec:efficiency}
We first evaluate decoding throughput across context lengths and batch sizes, 
followed by prefilling latency analysis, and lastly stress test the 
end-to-end request latency and throughput of all systems under different loads. 
vLLM is not suitable for direct comparison in decoding throughput, as it 
employs internal request scheduling policy~\cite{agrawal2023sarathi}, 
making it challenging 
to explicitly control the batch size. Thus we only include it in end-to-end 
evaluation. 

\begin{figure}[t!]
    \centering
    \includegraphics[width=\columnwidth]{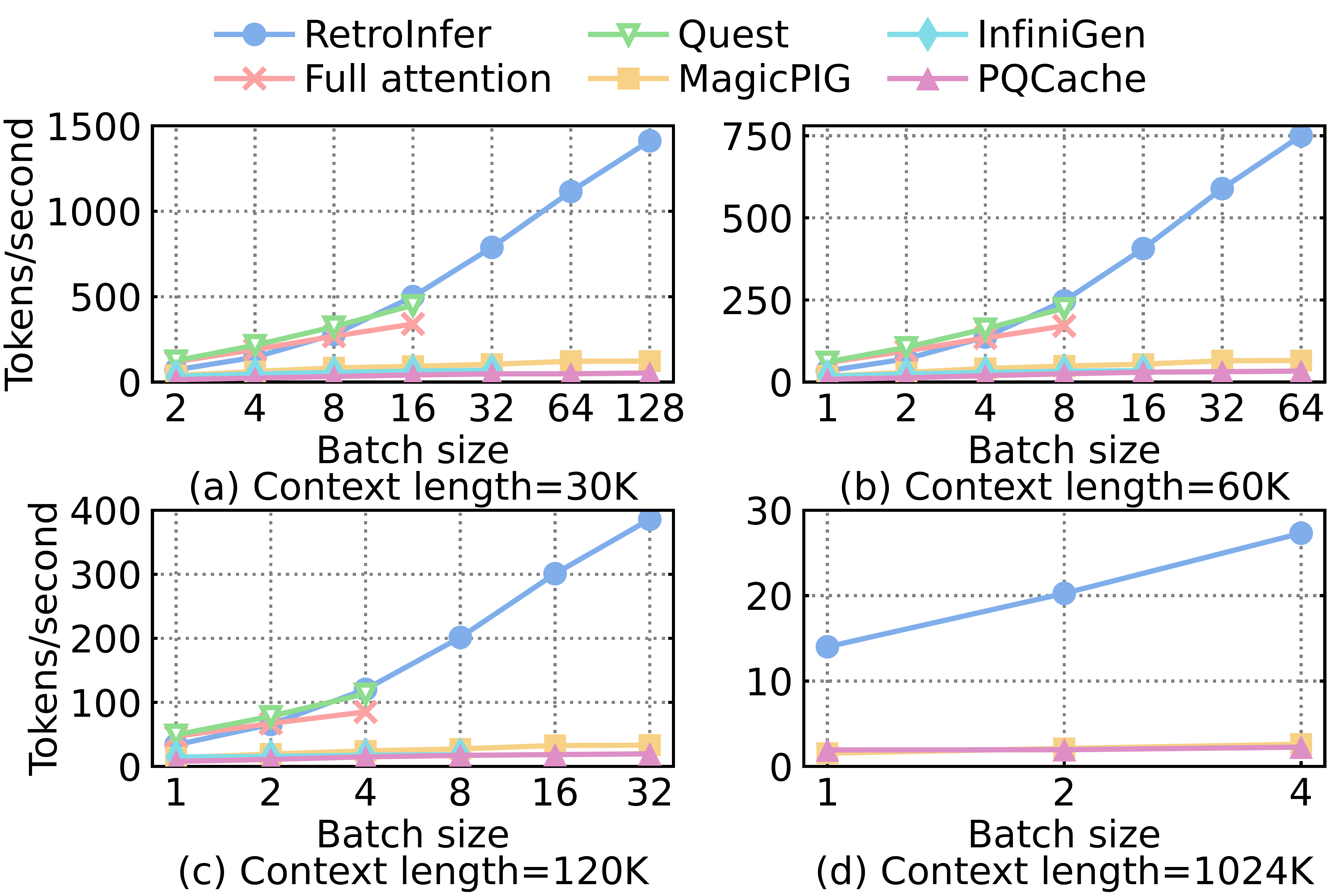}
    \caption{Decoding throughput at different context lengths (Llama3-8B-1048K). \newsysname{} scales well with the batch size and achieves the best throughput on all lengths.}
    \label{fig:over_tp}
\end{figure}

\minisec{Decoding Throughput}
Figure~\ref{fig:over_tp} shows the decoding throughput of \newsysname{} and 
other systems across four different context lengths ranging from 30K to 1M. 
For the context length of 
30K, 60K and 120K shown in Figure~\ref{fig:over_tp} (a--c), \newsysname{} 
outperforms full attention by 4.1$\times$, 4.4$\times$, and 4.4$\times$ respectively. 
The speedup over sparsity-based systems ranges from 3.3$\times$ to 22.8$\times$. 
The efficiency of \newsysname{} stems from 
wave buffer, significantly extending the batch sizes while maintaining low overhead. 
When the batch size is small, full attention and Quest 
show comparable or slightly higher throughput than \newsysname{} due to their
fully GPU processing, but they cannot scale beyond GPU memory.
MagicPIG's throughput is constrained by limited CPU compute, while InfiniGen 
suffers from low throughput due to relatively expensive speculative operations 
and frequent PCIe data transfers. 
PQCache falls behind mainly due to its inefficient GPU-CPU memory management 
and increasing overhead of fetching PQ codebook as the context length increases.

At 1M context, full attention and Quest incur OOM. 
InfiniGen faces the same issue as the key cache retained on GPU 
for speculation exceeds GPU memory.
Therefore, we compare only with the remaining systems. 
\newsysname{} outperforms MagicPIG and PQCache by 10.5$\times$ and 12.2$\times$, 
respectively. 
This demonstrates \newsysname{}'s efficiency for supporting extremely 
long-context. 

\begin{figure}[t!]
    \centering
    \includegraphics[width=\columnwidth]{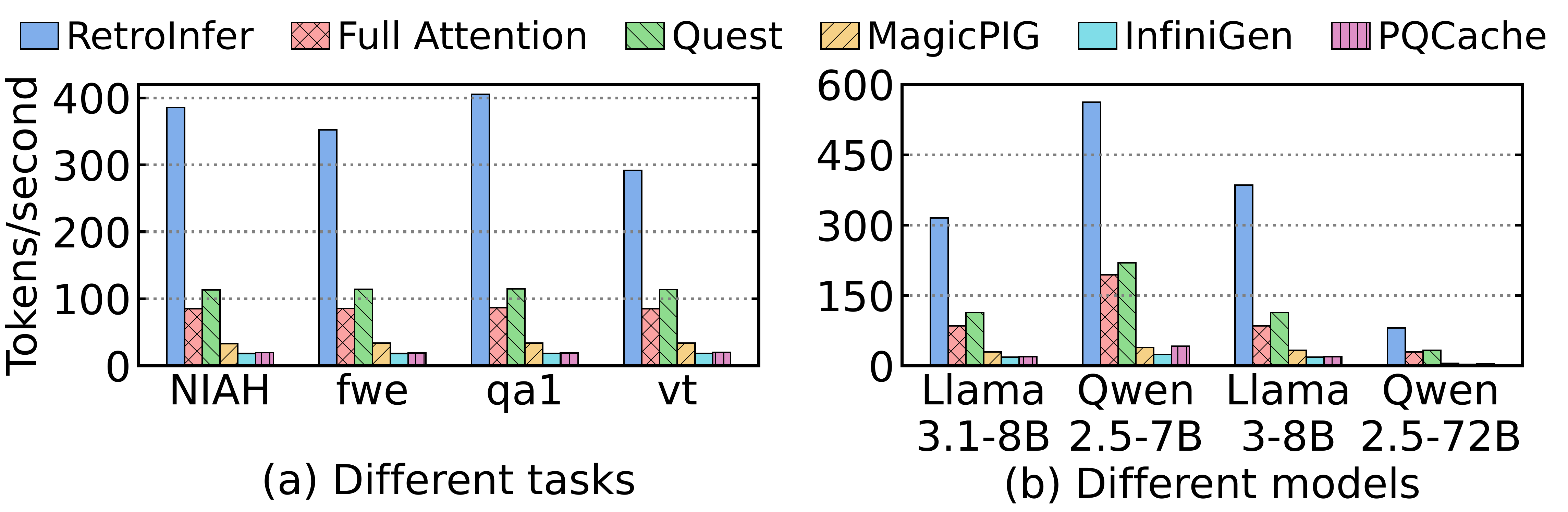}
    \caption{Maximum decoding throughput across tasks and models. \newsysname{} consistently outperforms others.}
    \label{fig:tp_task_model}
\end{figure}

We also evaluate the decoding throughput on different tasks and models. 
As shown in Figure~\ref{fig:tp_task_model}(a), \newsysname{} outperforms 
full attention by 3.4--4.6$\times$ across different tasks.
The throughput variation across tasks is due to differing cache hit ratios. 
The advantage over sparsity-based systems still holds, with 
\newsysname{} achieving 2.6--3.5$\times$ higher 
throughput than the best-performing baseline Quest on four tasks. 
Figure~\ref{fig:tp_task_model}(b) shows
that \newsysname{} outperforms baselines by 2.4--24.4$\times$ 
on four models. 
Thus, \newsysname{} is effective across 
different model architectures and model sizes (7B to 72B).
The 72B model scales well across multi-GPUs, maintaining an advantage 
ranging from 2.5$\times$ to 24.4$\times$ over baselines.

\begin{figure}[t!]
    \centering
    \begin{minipage}{0.44\columnwidth}
        \centering
        \includegraphics[width=.9\textwidth]{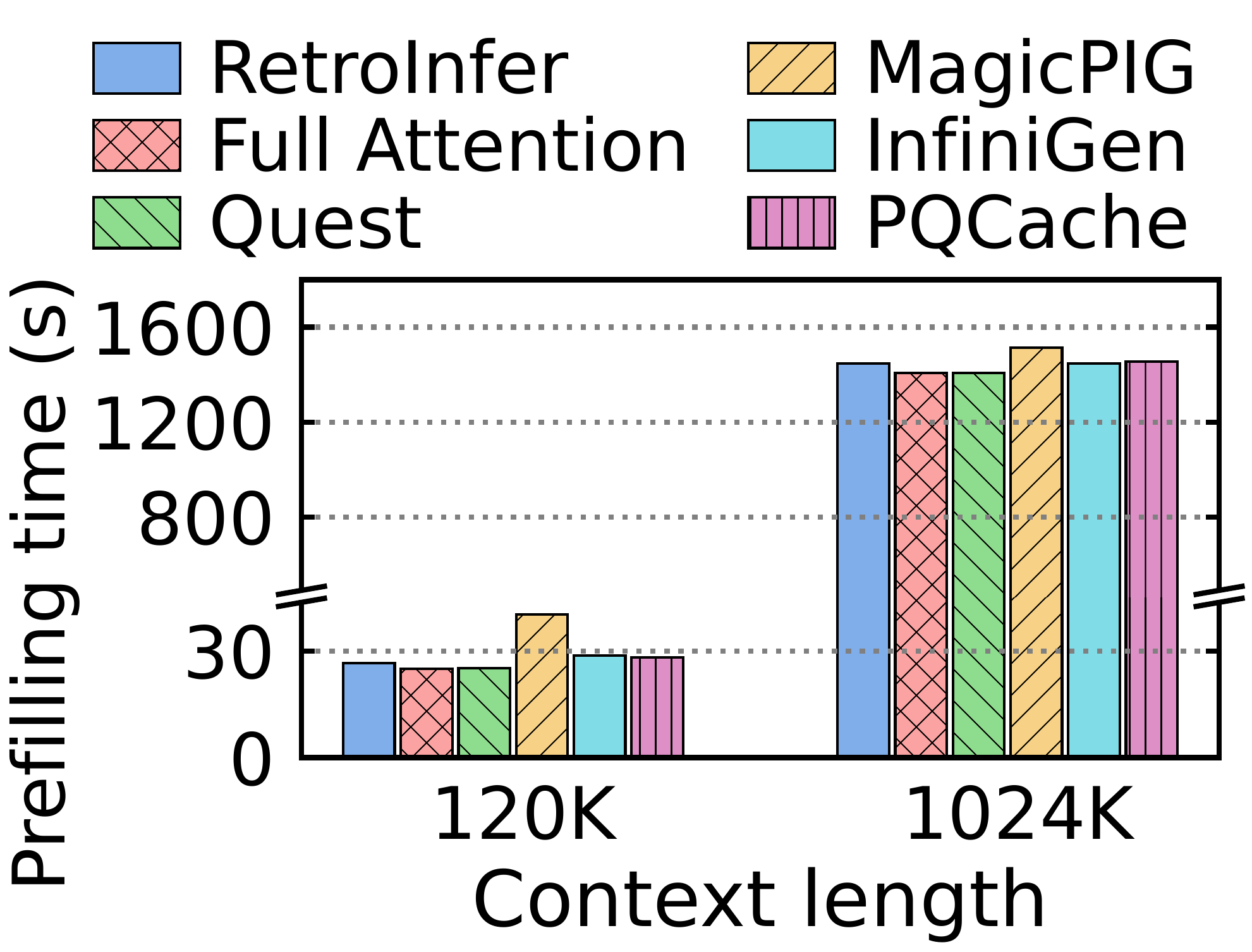}
        \caption{Prefilling latency under different context lengths. \newsysname's prefilling latency is only slightly higher than full attention due to lightweight index building.}
        \label{fig:prefill_latency}
    \end{minipage}
    \hfill
    \begin{minipage}{0.5\columnwidth}
        \centering
        \includegraphics[width=.93\textwidth]{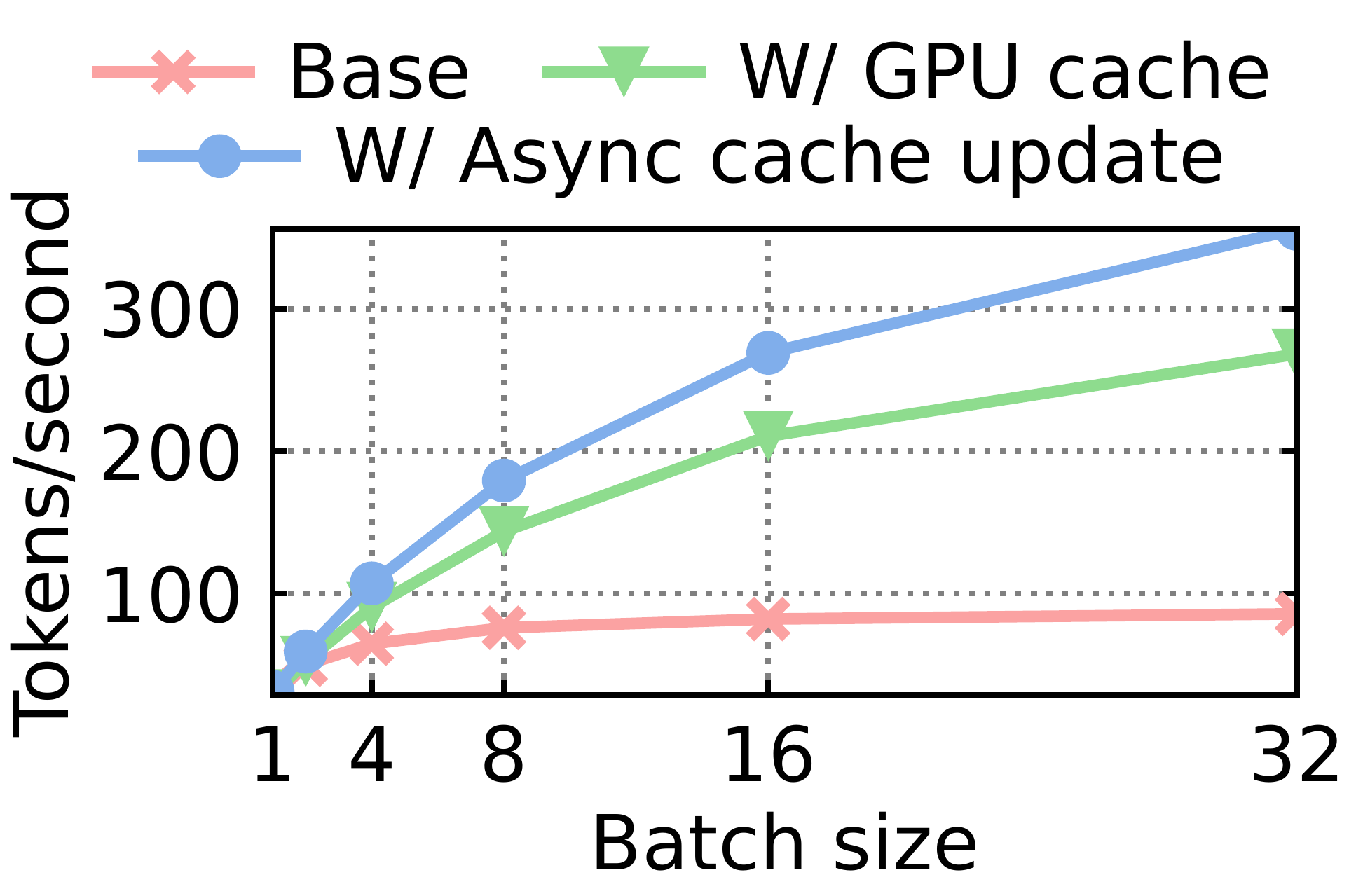}
        \caption{Effect of buffer design. ``Base'' offloads KV cache to CPU (no GPU cache), showing low scalability; ``W/ GPU cache'' improves throughput; ``W/ Async cache update'' further boosts the speed.}
        \label{fig:ablation_study}
    \end{minipage}
\end{figure}

\minisec{Prefilling Latency} 
\label{subsec:prefill_latency}
We evaluate the prefilling latency across context lengths. 
To avoid OOM errors at 1024K context, we offload the KV cache to CPU memory 
and measure the prefilling latency.
Offloading introduces only 0.4\% overhead for the prefilling.
As illustrated in Figure~\ref{fig:prefill_latency}, 
\newsysname{} achieves only 6\% and 3\% higher prefilling latency than 
full attention at 120K and 1M context lengths, respectively.
We attribute this to the lightweight segmented clustering and asynchronous 
wave buffer construction.
To reduce prefilling latency,
\newsysname{} can be combined with sparse prefilling methods; for example, at 1M context, XAttention and MInference help reduce the prefilling time to about $600 s$ and $220 s$, respectively.

\begin{figure}[t!]
    \centering
    \includegraphics[width=\linewidth]{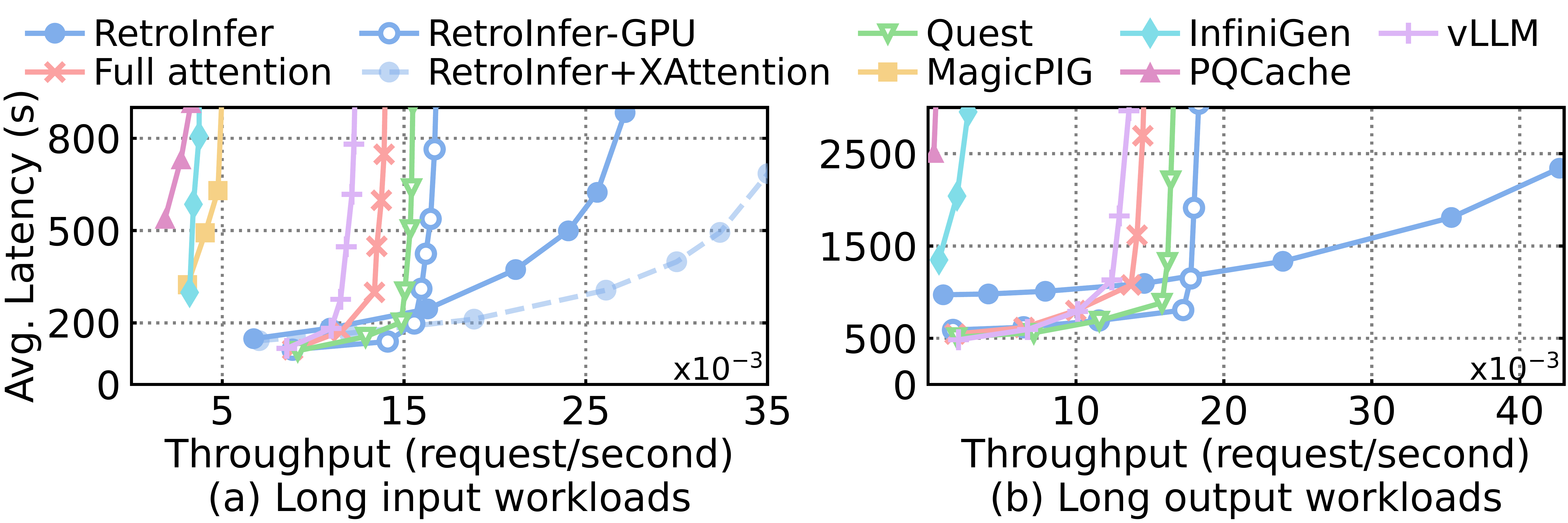}
    \caption{End-to-end request throughput vs. average latency at two workloads. \newsysname scales well on both.}
    \label{fig:e2e}
\end{figure}

\minisec{End-to-end Efficiency}
\label{subsec:e2e}
We further evaluate end-to-end request latency and throughput 
including both prefilling and decoding phases under varying loads.
We choose two representative workloads: (1) long input, consisting of 
120K input tokens and 4K output tokens; and (2) long output, 
consisting of 512 input tokens and 32K output tokens. 
Figure~\ref{fig:e2e} shows the latency-throughput curves.

For long-input workloads, under low load (e.g., two requests), 
\newsysname{} is slightly slower than GPU-only full attention, 
Quest, and vLLM because they do not incur PCIe transfers.
To better utilize GPU resources under lighter loads, we implement 
a GPU-only version of \newsysname{} denoted as ``\newsysname{}-GPU'', 
which achieves latency comparable to Quest.
When the load increases, \newsysname{} scales well and achieves 1.8--7.8$\times$ higher throughput than baselines at the same latency.
We further integrate a sparse prefilling technique XAttention with all baselines 
for lower prefilling latency, but only show ``\newsysname{}$+$XAttention'' 
in Figure~\ref{fig:e2e} for better readability.
With XAttention, \newsysname{} achieves even greater gains, delivering 2.1--10.5$\times$ throughput speedup. 

For long-output workloads, we exclude MagicPIG due to lack of index update support. 
As shown in Figure~\ref{fig:e2e}(b),
with the relatively small context ($\leq$ 32K),
the KV cache can fit into GPU memory under low loads.
At this load, \newsysname does not outperform GPU-only baselines, 
but switching to \newsysname{}-GPU achieves latency comparable to Quest. 
As load increases, \newsysname{} outperforms 
baselines by 2.7--70.8$\times$ by supporting larger batch sizes.

\subsection{Micro Analysis}
\label{subsec:micro}
We quantify the impact of individual design decisions of \newsysname{}
including different components of wave buffer, impact of different zone sizes and segment sizes and effect of attention estimation. 

\minisec{Effect of Design Decisions in Wave Buffer}
We assess the impact of two design choices in the wave buffer: GPU caching and asynchronous cache updates.
As shown in Figure~\ref{fig:ablation_study}, without GPU caching (Base), throughput 
cannot scale with increasing batch sizes, primarily due to PCIe bandwidth constraints. 
Adding GPU caching substantially reduces data transfer, enabling scalable throughput.
Furthermore, the asynchronous update of the mapping table further boosts throughput by
reducing latency on the critical path.

\begin{figure*}[t!]
    \centering
    \includegraphics[width=\textwidth]{figures/llama31_diff_zones.pdf}
    \caption{Impact of three zone sizes on maximum decoding throughput and task accuracy (Llama3.1-8B, 128K context).}
    \label{fig:llama31_diff_zones}
\end{figure*}

\begin{figure}[t!]
    \centering
    \includegraphics[width=0.98\linewidth]{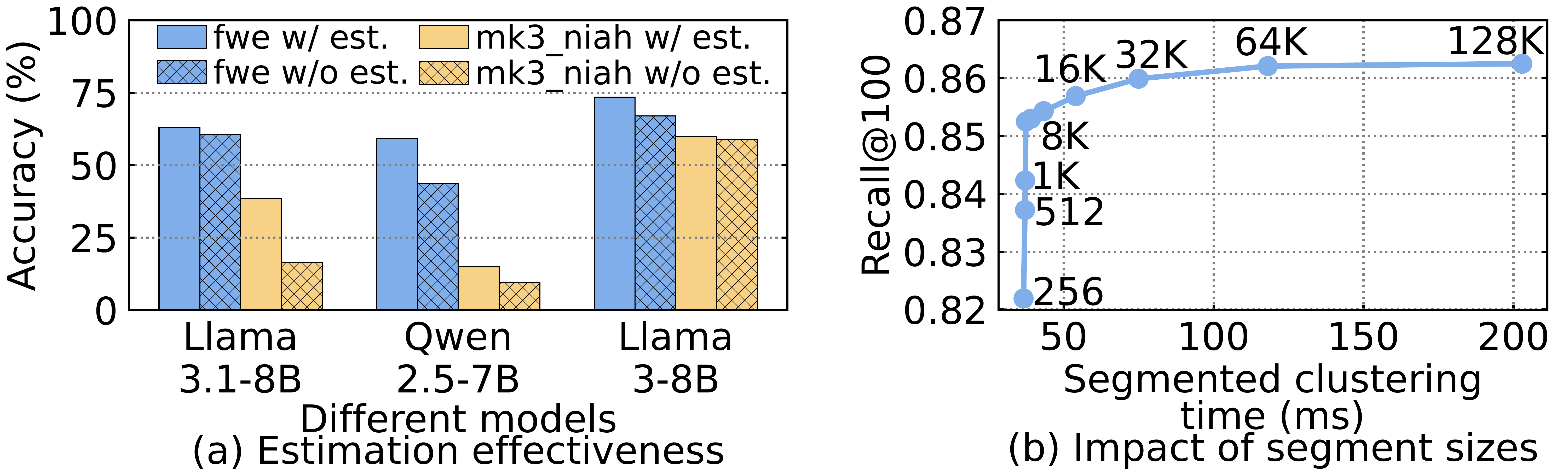}
    \caption{(a) Estimation helps improve the model accuracy. 
    (b) Index build time vs. accuracy at 128K context. The number next to the circle indicates the segment size. Segment size = 8K well balances index build time and clustering accuracy.}
    \label{fig:index_time_recall_and_est_ablation}
\end{figure}

\minisec{Impact of Different Zone Sizes}
We study the impact of different zone sizes on model accuracy 
and efficiency using two representative tasks from RULER. 
The retrieval task \textit{s3\_niah} exhibits high sparsity, whereas \textit{qa\_1} shows 
more sparsity variation depending on the questions posed to the context.
When varying the size of one zone, we keep other zone sizes fixed as in the evaluation setting. We report the decoding throughput on A100 and A6000 GPUs (48 GB memory)~\cite{A6000} respectively to evaluate efficiency across hardware.

As shown in Figure~\ref{fig:llama31_diff_zones}(a-b), increasing the retrieval budget improves task accuracy but greatly reduces throughput due to increasing PCIe data transfers. \newsysname{} reaches full-attention accuracy at a 1.8\% retrieval budget across tasks, with diminishing returns beyond this point. The 1.8\% budget is robust across tasks due to \newsysname{}'s estimation zone, which efficiently captures varying sparsity. We analyze this in detail below.

Figure~\ref{fig:llama31_diff_zones}(c-d) shows that increasing estimation budget improves accuracy, reaching full-attention accuracy at $\sim$23.2\%, with greater gains for tasks with higher sparsity, like \textit{qa\_1}. 
Unlike retrieval, estimation imposes lower
overhead on decoding throughput thanks to its lower computational cost 
and avoidance of PCIe transfers. 
This motivates our choice to keep the retrieval zone small (1.8\%) 
and allocate a relatively larger estimation zone (23.2\%) to capture varying
sparsity patterns, for both high throughput and accuracy.

We empirically set the steady zone size to $4+64$ based on common practice~\cite{magicpig,streamingllm}. 
More configurations (e.g., sink tokens only) are evaluated in Figure~\ref{fig:llama31_diff_zones}(e-f).  
Both sink and local window tokens are important, with sink tokens contributing more to accuracy. Larger steady zone sizes yield marginal accuracy gains, making $4+64$ sufficient. 
Moreover, steady zone size has minimal impact on throughput due to its relatively small size.

\minisec{Impact of Attention Estimation}
We quantify its effect by comparing \newsysname{} with and without attention estimation. As depicted in Figure~\ref{fig:index_time_recall_and_est_ablation}(a), estimation enhances task accuracy by up to 20\%. Importantly, this improvement is achieved without compromising efficiency, as it can be overlapped with wave buffer accesses.

\minisec{Impact of Different Segment Sizes}
\label{sec:segment_exp}
We study the impact of different segment sizes 
on both clustering execution time and 
index accuracy, using KV data from layer 10 of Llama3-8B-1048K with \textit{mv\_niah} task under 128K context. Recall@100 is used as the metric for clustering accuracy. 
As shown in Figure~\ref{fig:index_time_recall_and_est_ablation}(b), 
reducing segment size from 128K (standard global $k$-means) to 8K results in less than 1\% drop in recall, benefiting from spatial locality, while reducing index build time by $80\%$. 
However, further decreasing the segment size worsens clustering quality. 
Experiments across models and tasks show consistent trends. Based on this, we set the segment size to 8K as a balanced configuration.
\section{Related Work}
\label{sec:related-work}

\minisec{LLM Serving Systems}
Recent research has explored various strategies for LLM serving 
efficiency~\cite{yu2022orca, kwon2023efficient, zhong2024distserve, agrawal2024taming, sun2024llumnix, 
sheng2024fairness, lin2024parrot, fu2024serverlessllm, zheng2024sglang, Gao_2025}.
LoongServe~\cite{wu2024loongserve} proposes an elastic sequence parallelism 
scheduling for long-context requests 
while Helix~\cite{mei2025helix} focuses on scheduling requests across heterogeneous GPU clusters.
They are complementary to \newsysname{}, which focuses on efficiently 
managing long-context KV cache.
Many works~\cite{gao2024cost, qin2025mooncake, gao2025fast, chen2025impress, yu2025stateful} 
offload KV cache to external storage for reuse across requests. 
\newsysname{} can be combined with these works 
to reduce prefilling latency.

\minisec{Sparse Attention} 
To accelerate long-context inference, many works exploit attention sparsity to reduce 
computation and memory pressure. 
Static sparsity methods~\cite{li2024snapkv, streamingllm, xiao2024duoattention} use 
fixed patterns during decoding but compromise model accuracy. 
This motivates dynamic sparsity systems~\cite{xiao2024infllm, h2o, quest, yang2025lserve, lee2024InfiniGen, 
magicpig} to heuristically select important tokens per step for better accuracy.
Some works~\cite{liu2024retrievalattention, hooper2024squeezed, alayadb} 
explore vector indexes for KV cache retrieval but restrict their use to prefix caching or offline inference due to the high index construction cost. 
ClusterKV~\cite{liu2024clusterkv} employs clustering but lacks adaptability to varying sparsity, leading to suboptimal performance. 
Different from these works, \newsysname{} designs wave index and wave buffer holistically for efficient KV cache management.

PyramidKV~\cite{Pyramid} and D2O~\cite{wan2024d2o} adapt retrieval budgets 
across layers but ignore sparsity variation across queries and tasks.
Tactic~\cite{Tactic} adjusts budgets via distribution fitting, 
though its accuracy depends on fitting precision. Our work focuses more 
on system-level design than on pure algorithmic improvements. 
Some sparse attention approaches~\cite{gao2024seerattention, yuan2025native} require 
model retraining to learn dynamic sparsity, while \newsysname{} is training-free.
Other techniques, including KV cache quantization~\cite{liu2024kivi, hooper2024kvquant, 
zhang2024kv, zhang2024unifying} and 
prefill sparsity acceleration~\cite{jiang2024minference, xuxattention}, are 
complementary to our work.

\minisec{LLM Weight Sparsity} 
Model weight compression or offloading~\cite{lin2024awq, zhao2024atom, alizadeh2024llm, 
sheng2023flexgen, song2024powerinfer, xue2024powerinfer} 
reduces the memory requirement for the LLM itself. 
They are orthogonal to \newsysname{}, which targets long-context scenarios 
where the KV cache is the memory bottleneck.

\minisec{Vector Data Management}
Managing vector data is an important problem in data management domains, 
including vector database systems~\cite{yang2020pase, wang2021milvus, guo2022manu, chen2024singlestore, pan2024survey}, 
vector indexing and retrieval techniques~\cite{fu2017fast, malkov2018efficient, wang2020deltapq, Lu2021hvs, 
mohoney2023high, zhao2023towards, wei2024detlsh, patel2024acorn, ootomo2024cagra}. 
Retrieval-augmented generation (RAG)~\cite{lewis2020retrieval, zhao2024chat2data, 
liu2025tigervector} retrieves external knowledge to enhance LLMs, which is orthogonal to KV cache management in 
\newsysname{} that focuses on the vector management inside the LLM context.

\section{Conclusion}
\label{sec:conclusion}

We presented \newsysname{}, a vector storage engine for high throughput
inference on long-context LLMs. To efficiently manage the KV cache and exploit
attention sparsity, \newsysname{} introduces a co-designed software stack
integrating an attention-aware vector index (\textit{wave index}) with a runtime
control plane (\textit{wave buffer}) for heterogeneous memory systems. The wave
index improves the trade-off between attention accuracy
and retrieval cost, while the wave buffer decouples attention approximation from system efficiency, enabling scalable execution across GPUs and CPUs. We demonstrate that \newsysname{} not only achieves significant speedups over baselines,
but also preserves model accuracy.

\begin{acks}
 We thank all anonymous reviewers for their insightful comments.
 This work is supported in part by the National Key R\&D Program of China under Grant No. 2024YFB4505701.
\end{acks}

\balance
\bibliographystyle{ACM-Reference-Format}
\bibliography{ref}

\end{document}